\documentclass{article}

\usepackage{booktabs}
\usepackage{times}
\usepackage{paralist}
\usepackage{latexsym}
\usepackage{graphicx}
\usepackage{multirow}
\usepackage{subfig}
\usepackage{amsthm}
\usepackage{enumitem}
\usepackage{todonotes}
\usepackage{xspace}
\usepackage[export]{adjustbox}
\usepackage{amsmath,amssymb}
\DeclareMathOperator{\E}{\mathbb{E}}
\DeclareMathOperator{\Var}{\text{Var}}
\DeclareMathOperator{\ELBO}{\mathrm{ELBO}}
\DeclareMathOperator{\KL}{\text{KL}}
\DeclareMathOperator{\HH}{\mathcal{H}}

\DeclareMathOperator{\LL}{\mathcal{L}}

\newtheorem{property}{Property}

\newcommand \NOTE[1]{{\bf \color{red} \\NOTE:#1.\\}}

\newcommand{\methodshort}{{DGM-VAE}\xspace} 
\newcommand{\dgmvae}{{DGM-VAE}\xspace} 
\newcommand{\dcmvae}{{DCM-VAE}\xspace} 
\newcommand{\regc}{\mathcal{R}_c}
\newcommand{\regz}{\mathcal{R}_z}

\usepackage{microtype}
\usepackage{booktabs} 

\usepackage{hyperref}

\usepackage[accepted]{icml2020}

\begin{document}

\twocolumn[
\icmltitle{Dispersed Exponential Family Mixture VAEs for Interpretable Text Generation}





\begin{icmlauthorlist}
\icmlauthor{Wenxian Shi}{bd}
\icmlauthor{Hao Zhou}{bd}
\icmlauthor{Ning Miao}{bd}
\icmlauthor{Lei Li}{bd}
\end{icmlauthorlist}

\icmlaffiliation{bd}{ByteDance AI lab}

\icmlcorrespondingauthor{Wenxian Shi}{shiwenxian@bytedance.com}
\icmlcorrespondingauthor{Hao Zhou}{zhouhao.nlp@bytedance.com}
\icmlcorrespondingauthor{Ning Miao}{miaoning@bytedance.com}
\icmlcorrespondingauthor{Lei Li}{lileilab@bytedance.com}


\vskip 0.3in
]



\printAffiliationsAndNotice{}  

\begin{abstract}
Deep generative models are commonly used for generating images and text. 
Interpretability of these models is one important pursuit, other than the generation quality.
Variational auto-encoder (VAE) with Gaussian distribution as prior has been successfully applied in text generation, but it is hard to interpret the meaning of the latent variable.
To enhance the controllability and interpretability, one can replace the Gaussian prior with a mixture of Gaussian distributions (GM-VAE), whose mixture components could be related to hidden semantic aspects of data.
In this paper, we generalize the practice and introduce DEM-VAE, a class of models for text generation using VAEs with a mixture distribution of exponential family.
Unfortunately, a standard variational training algorithm fails due to the \emph{mode-collapse} problem.
We theoretically identify the root cause of the problem and propose an effective algorithm to train DEM-VAE.
Our method penalizes the training with an extra \emph{dispersion term} to induce a well-structured latent space. 
Experimental results show that our approach does obtain a meaningful space, and it outperforms strong baselines in text generation benchmarks.
The code is available at \url{https://github.com/wenxianxian/demvae}.
\end{abstract}

\section{Introduction}
\label{sec:intro}
Text generation is one of the most challenging problems in Natural Language Processing~(NLP), which is widely applied to tasks such as machine translation~\citep{brown1993mathematics} and dialog system~\citep{young2013pomdp}. Though generation quality is the main concern of most work, the interpretability for generation process is also of great importance. Interpretable generation models could explore the latent data structures, such as topics~\citep{wang2019topic} and dialog actions~\citep{zhao2018unsupervised}, and uses them to guide text classification and further generation.
Among deep generative models, variational auto-encoder~\citep[VAE]{kingma2013auto, rezende2014stochastic} is especially suitable for interpretable text generation because it maps sentences to a regularized latent variable $z$, which can be used to derive interpretable structures~\citep{bowman2016generating, miao2016neural, semeniuta2017hybrid, xu2018spherical}.

However, the continuous latent variable $z$ of vanilla VAE makes it difficult to interpret discrete attributes, such as topics and dialog actions.
Recently, \citet{zhao2018unsupervised} propose to replace the continuous latent variable with a discrete one for better interpretation in generating dialog, where the discrete latent variable represents the dialog actions in their system, producing promising results even in an unsupervised setting. 
Unfortunately, the expressiveness of VAE with only a discrete latent variable $c$ is very limited. It only contains $\log(\#c)$ bits of information, where $\#c$ is the number of all possible values of $c$. So it is impossible for discrete VAE to express the complicated sentence space. 

\begin{figure}[tp]
\footnotesize
    \subfloat[GM-VAE]{
     \label{fig:collape_gmvae}
     \centering
     \includegraphics[width=0.43\linewidth]{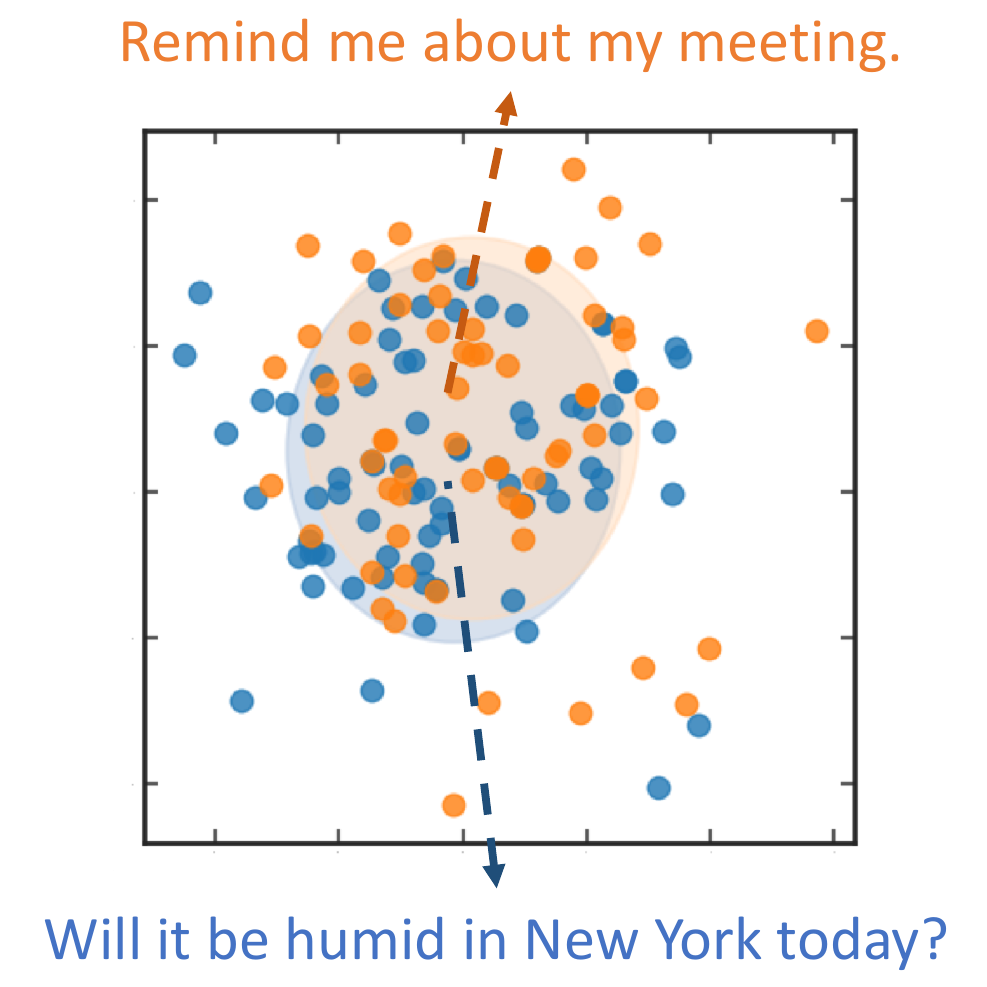} 
      \hspace {10 pt}
     }
     \subfloat[\methodshort]{
     \label{fig:collapse_bgmvae}
     \centering
     \includegraphics[width=0.43\linewidth]{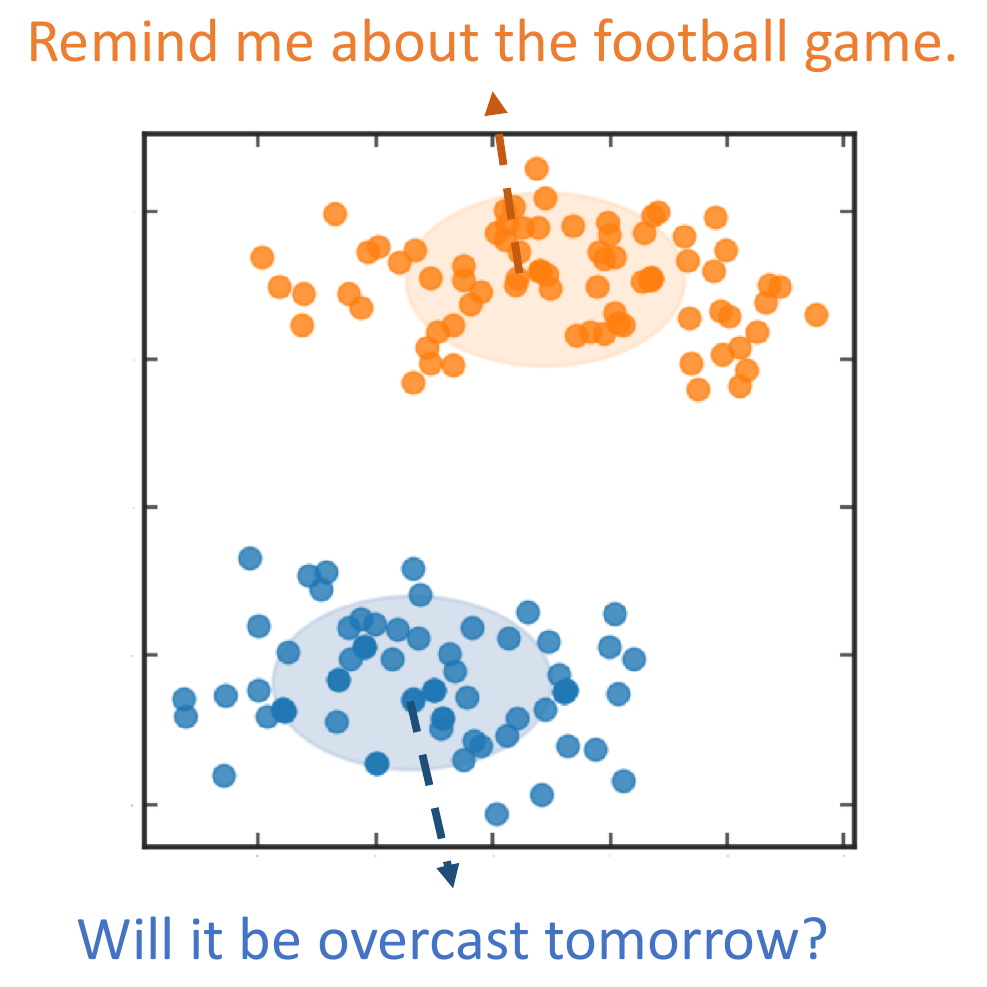} 
      }
      \vspace{-6 pt}
    \caption{Latent space learned by GM-VAE and our proposed DEM-VAE with a specific Gaussian mixture prior (named \methodshort) for dialog intention recognition. Notice that our proposed \methodshort avoids mode-collapse of GM-VAE. As a result, requests with different intentions are clearly separated into two clusters. } \label{fig:all_collapse}
\end{figure}

To solve the above concern, Gaussian mixture VAE (\textsc{GM-VAE}) offers a natural choice, which enjoys the benefits of both discrete and continuous latent space. 
Each component $c$ represents a discrete attribute, while continuous latent variable $z$ in each component represents different sentences with the same attribute.
However, \textsc{GM-VAE} is prone to \emph{mode-collapse}, which makes it difficult to train. In other words, different components tend to have very close means and variances after training, which makes GM-VAE degenerate to vanilla VAE with only one Gaussian component~(Fig.~\ref{fig:all_collapse}).
As a result, GM-VAE fails to capture the multi-modal data structure, and cannot effectively utilize the discrete latent variables. For example, as illustrated in Fig.~\ref{fig:collape_gmvae}, utterances ask about the weather and requesting appointments are mapped into the same mode.  
In this paper, we theoretically demonstrate that mode-collapse does not only occur in GM-VAE, which is a general problem for VAEs with exponential family mixture priors~(EM-VAE). 
We find that the problem is intrinsically caused by a ``dispersion'' term in the evidence lower bound~($\ELBO$), which makes it extremely easy to fall into the solution of mode collapse.

To address the mode-collapse problem, we propose Dispersed EM-VAE~(DEM-VAE), which introduces an extra \emph{dispersion} term in the training objective (Fig.~\ref{fig:collapse_bgmvae}) for overcoming the contraction tendency introduced by ELBO.
Experimental results show that our proposed DEM-VAE alleviates the mode-collapse problem effectively.
Furthermore, DEM-VAE outperforms strong baselines in interpretable text generation benchmarks.
Although DEM-VAE moderately decreases the sentence likehoood because of the inclusion of external training term, it outputs sentences with significantly better quality in aspects of other evaluation metrics such as rPPL and BLEU scores. 

\section{Related Work}
\label{sec:related}

\noindent\textbf{VAEs for Language Generation.}
Variational auto-encoders are proposed by \citet[VAEs]{kingma2013auto} and  \cite{rezende2014stochastic}, and applied by \citet{bowman2016generating} for natural language generation.
VAEs are extended by many following works in various specific language generation tasks, such as dialog generation \citep{serban2017a, wen2017latent, zhao2017learning, zhao2018unsupervised}, summarization~\citep{li2017deep} and other natural language generation tasks~\citep{miao2016neural,zhang2016variational, semeniuta2017hybrid, gupta2018a, xu2018spherical, ye2019variational, bao2019generating}.

Additionally, \citet{wen2017latent} and \citet{zhao2018unsupervised} propose to replace the continuous latent variable with a discrete one for interpretable sentence generation. 
\citet{kingma2014semi} propose the \textit{semi}-VAE for semi-supervised learning.
This model is then adopted by \citet{hu2017toward,zhou2017multi-space} for style-transfer and labeled sequence transduction, respectively.
Different from GM-VAE, continuous and discrete latent variables in \textit{semi}-VAE are independent.

\noindent\textbf{Gaussian Mixture VAEs.}
Using Gaussian mixture models as priors in VAEs is not new. 
Gaussian mixture variational auto-encoder has been used in the unsupervised clustering~\citep{dilokthanakul2016deep, jiang2017variational}, obtaining promising results.
\citet{wang2019topic} used GMM as priors for topic-guided text generation.
In this work, we apply GM-VAE for interpretable text generation and propose the \methodshort to address the mode-collapse problem according to our theoretical analysis.

\noindent\textbf{KL Collapse vs. Mode Collapse.} 
The vanilla VAE models usually suffer from the KL collapse problem~\citep{bowman2016generating} in language generation, in which the KL regularization term will quickly collapse to 0.
A line of following work~\citep{bowman2016generating, zhao2017learning, zhao2018unsupervised, higgins2017beta-vae, he2018lagging, li2019surprisingly} is proposed to avoid the KL collapse problem. 

These methods are different from ours in two-folds.
First, the problem to be solved is different. Our method aims to address the mode-collapse in the mixture of exponential family VAEs, while these methods try to fix the posterior collapse in general VAEs. Mode collapse and posterior collapse are not the same. Posterior collapse means that the posterior distributions of all latent variables collapse to their priors, while mode-collapse means the multiple modes of prior collapses to one mode. Mode-collapse might also occur when the posterior does not collapse.
Second, although our method and previous methods of solving KL collapse have similar solutions that find the problem-causing term in the objective and weakens it, a heuristic dispersion term instead of the whole KL term is introduced according to our theoretical analysis. The dispersion term is explicitly related to the difference between prior components and is specifically effective for the mode collapse problem.

\section{Approach}
\label{sec:method}
We first describe the VAEs with mixture of exponential family priors for text generation, and investigate the mode-collapse issue in them. 
Based on the results of the investigation, we propose dispersed exponential family mixture VAEs to fix this issue, among which dispersed Gaussian mixture VAE (\dgmvae) is particularly exemplified. 

\subsection{Mixture of Exponential Family VAEs}
\label{sec:gmvae}

\begin{figure}[tp]
\footnotesize
\centering
    \subfloat[VAE]{
     \label{fig:pgm_vae}
     \centering
     \includegraphics[scale=0.28]{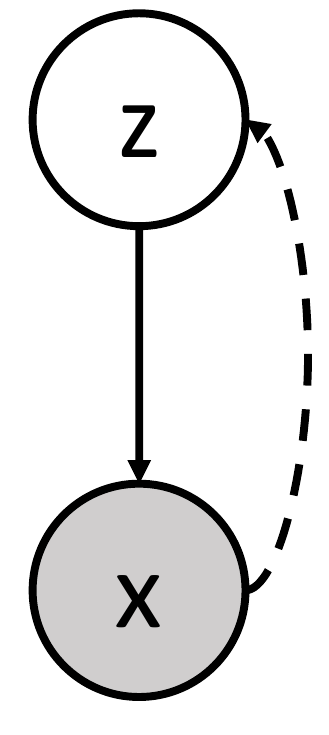}
      \hspace {3 pt}
     }
     \subfloat[DI-VAE]{
      \hspace {3 pt}
     \label{fig:pgm_divae}
     \centering
     \includegraphics[scale=0.28]{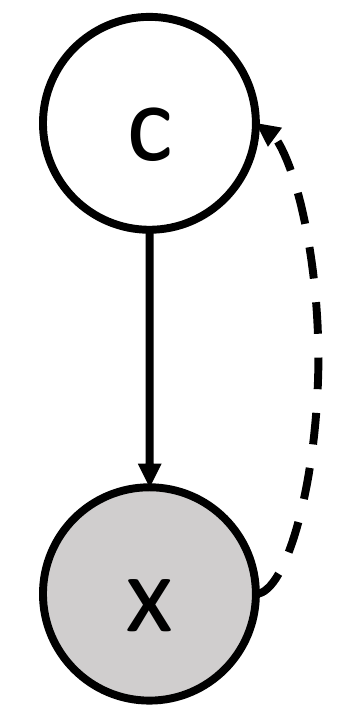}
       \hspace {3 pt}
     }
      \hspace {3 pt}
     \subfloat[semi-VAE]{
     \label{fig:pgm_semivae}
     \includegraphics[scale=0.28]{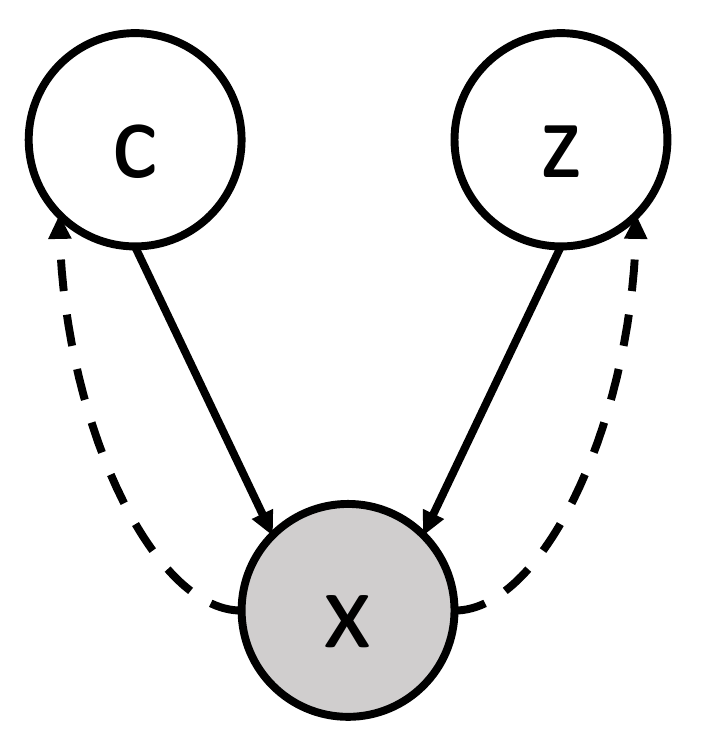}}
       \hspace {3 pt}
     \subfloat[GMVAE]{
     \label{fig:gmvae}
       \hspace {2 pt}
     \includegraphics[scale=0.28]{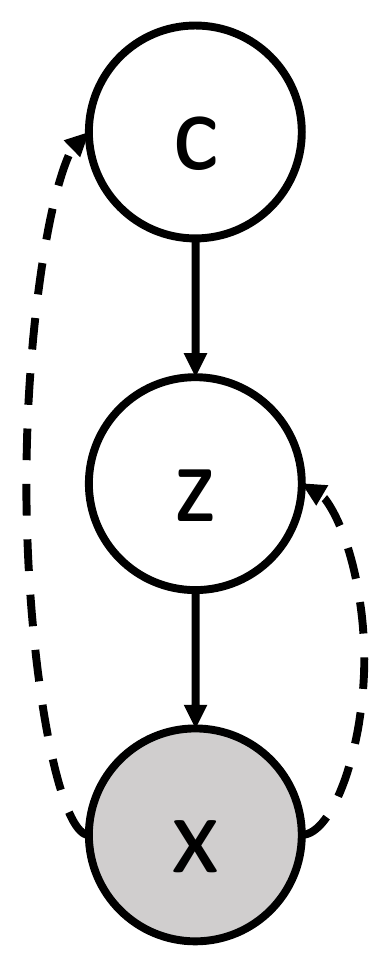}
      \hspace {2 pt}
     }
    \vspace{-6 pt}
    \caption{ Graphical models for various VAEs. 
     $z$ and $c$ are continuous and discrete latent variables, respectively. 
     $x$ is observed data. 
     The solid lines are conditional dependencies and dashed lines represent variational posteriors.}\label{fig:pgm}
\end{figure}

Mixture of Exponential Family VAEs are variational auto-encoders that adopts the mixture of exponential family distributions as its prior.
GM-VAE is the most popular exponential family mixture VAE, whose prior is a mixture of Gaussian~\citep{Bishop2006}.
GM-VAE employs a discrete latent variable $c$ to represent the mixture components, and a continuous latent variable $z$ dependent on $c$.
In this model, the marginal likelihood of a sentence $x$ is:
\begin{align} \label{eq:original_elbo}
    p(x) = \int \sum_c p_\eta(z, c) p_\theta(x|z)dz,
\end{align}
in which $\theta$ is the parameters of generation model which generates $x$ from $z$. $p_\eta(z, c)$ is the mixture \textit{prior} distribution with parameters $\eta$ and can be factorized by $p(z,c) = p(c)p_\eta(z|c)$.
Intuitively, $p(c)$ could be assumed as a uniform distribution; while $p_\eta(z|c)$ is an exponential family distribution, such as Gaussian, of the corresponding $c$-th component.

Fig.~\ref{fig:pgm} compares the graphical models of VAE variants. The vanilla VAE~\citep{kingma2013auto} uses a latent continuous variable, while the discrete VAE models \citep{zhao2018unsupervised} use a discrete one. The semi-VAE~\citep{kingma2014semi} combines independent discrete and continuous latent variables, and the GM-VAE~\citep{dilokthanakul2016deep, jiang2017variational} impose dependency between them. 

\noindent\textbf{Testing.} During testing, a mixture  component $c$ is first chosen according to the prior distribution $p(c)$. Then the variable $z$ is sampled from the chosen component $p_\eta(z|c)$. As in \citet{bowman2016generating}, a \textit{generation network} takes $z$ as input and generate the sentence $x$ through a decoder $p_\theta(x|z)$. 

\noindent\textbf{Training.} 
Optimizing and inference for Eq.~\ref{eq:original_elbo} is difficult. Following previous work of ~\citet{kingma2013auto} and \citet{rezende2014stochastic}, we use a variational \textit{posterior} distribution $q_\phi(z,c|x)$ with parameters $\phi$ to approximate the real posterior distribution $p(z, c|x)$.
With the mean field approximation~\citep{xing2002generalized}, $q_\phi(z, c|x)$ can be factorized as: $q_\phi(z, c|x) = q_\phi(z|x)q_\phi(c|x).$

The posterior $q_\phi(z|x)$ could be assumed as a multivariate Gaussian distribution, whose mean $\mu_\phi(x)$ and variance $\sigma^2_\phi(x)$ are obtained through a neural network (\textit{recognition network}).
$q_\phi(c|x)$ could be implemented by a neural network classifier.

Instead of optimizing the marginal likelihood in Eq.~\ref{eq:original_elbo}, we maximize a evidence lower bound~($\ELBO$). The $\ELBO$ can be decomposed as the summation of a reconstruction term and regularization terms for $c$ and $z$, respectively:
\begin{equation*}\label{eq:ELBO}
\begin{split}
    \ELBO &= \E_{q_\phi(z,c|x)}\big{(}\log p_{\theta, \eta}(x,c,z) - \log q_\phi(z, c|x) \big{)}\\
    &=\underbrace{\E_{q_\phi(z|x)q_\phi(c|x)}\log p_\theta(x|z)}_\text{Reconstruction gain ($\mathcal{R}_\text{rec}$)}  \\
    & +\underbrace{\E_{q_\phi(z|x)q_\phi(c|x)} \big{(} \log p(c) - \log q_\phi(c|x) \big{)}} _\text{Regularization on c ($\mathcal{R}_c$)} \\ & +\underbrace{\E_{q_\phi(z|x)q_\phi(c|x)} \big{(} \log p_\eta(z|c) - \log q_\phi(z|x) \big{)}.} _\text{Regularization on z ($\mathcal{R}_z$)} 
\end{split}
\end{equation*}
All parameters including $\theta$, $\phi$ and $\eta$ could be jointly trained with reparameterization tricks~\citep{kingma2013auto, jang2016categorical} 

However, exponential family mixture VAEs often encounter \textit{mode-collapse}, where all components of mixture prior degenerate to one distribution.

\subsection{Mode-Collapse Problem}
\label{sec:meancollapse}
We further investigate the $\ELBO$ objective function to analyze mode-collapse. To this end, we present that the regularization terms of exponential family mixture VAE's $\ELBO$, $\regc$ and $\regz$, are responsible for the mode collapse problem. 
We only give explanations and remarks of the proof, with the details included in the supplementary materials.

According to the definition of exponential family, we represent the probability density function of  $c$-th component in mixture prior by the natural parameters
\begin{equation}\label{eq:natural_paramter}
    p_{\eta}(z|c) = \exp (<{\boldsymbol \eta_c}, {\boldsymbol \phi(z)}> - A(\boldsymbol \eta_c)),
\end{equation}
in which $\boldsymbol \phi(z)$ is a vector of functions named sufficient statistics, and $\boldsymbol \eta_c$ is the corresponding parameters vector.  $<\boldsymbol \eta_c, \boldsymbol \phi(x)>$ means the inner-product of vector $\boldsymbol \eta_c$ and $\boldsymbol \phi$. $A(\boldsymbol \eta_c)$ is the log-partition function for normalizing the probability density. Taking Gaussian distribution as an example, $\boldsymbol \phi(z)=[z,z^2]$, $\boldsymbol \eta_c = [\eta_1, \eta_2] = [\frac{\mu_c}{\sigma^2_c}, -\frac{1}{2\sigma^2_c}]$ and $A(\eta_1, \eta_2)=-\frac{\eta_1^2}{4\eta_2}-\frac{1}{2}\log (-2\eta_2)$.

Plugging the definition of Eq.~\ref{eq:natural_paramter} in regularization $\regz$, $\regz$ could be re-written as two terms: an ``average'' KL regularization term and a ``dispersion'' term:
\begin{equation*}
\begin{split} \label{eq:proof}
    \E_{q_\phi(z|x)q_\phi(c|x)} & \log \frac{p_\eta(z|c)}{q_\phi(z|x)}  = \underbrace{-\KL(q_\phi(z|x)||\hat{p}_{\overline{\eta}}(z|x)))}_\text{Average $\mathcal{R}_z$} \\
   &- \underbrace{(\E_{q_\phi(c|x)}A(\boldsymbol \eta_c) - A(\E_{q_\phi(c|x)} \boldsymbol \eta_c))}_\text{Dispersion term $\LL_d$}.
\end{split}
\end{equation*}
The ``average'' KL regularization term is a KL divergence between posterior $q_\phi(z|x)$ and an ``average'' prior: $\hat{p}_{\overline{\eta}}(z|x)=\exp \left( \left< \overline{\boldsymbol \eta},  \boldsymbol\phi(x) \right> - A( \overline{\boldsymbol\eta}) \right)$, which is an exponential family distribution parameterized by ``average'' parameters $ \overline{\boldsymbol \eta} = \E_{q_\phi(c|x)}\boldsymbol \eta_c$.
Note that for exponential families, the domain of the parameters $\boldsymbol \eta$ is an convex set, thus the  $ \overline{\boldsymbol \eta}$ is feasible parameters.

The ``dispersion'' term is respect to priors parameters and $q_\phi(c|x)$. In the following, we demonstrate two properties of the dispersion term  inducing the mode-collapse.
\begin{property} 
\label{property-1}
Dispersion term $\LL_d=\E_{q_\phi(c|x)}A(\boldsymbol \eta_c) - A(\E_{q_\phi(c|x)} \boldsymbol \eta_c) \geq 0$ and it gets to zero when parameters $\boldsymbol \eta_c$ of different components $c$ collapse to the same one, when $q_\phi(c|x)$ does not have one-hot probability mass.
\end{property}

For minimal representation of exponential family\footnote{Sufficient statistics are linearly independent to each other.}, the Property~\ref{property-1} could be easily proven by the strictly convexity of log-partition function~\citep{wainwright2008graphical}. Besides, the $\regc$ prevent the  $q_\phi(c|x)$ to be one-hot because the $p(c)$ is always assumed to be uniform.  

\begin{property} 
\label{property-2}
Minimizing the dispersion term $\LL_d$  makes the weighted variance of prior parameters, defined as
\begin{equation*}\label{eq:var}
\small
\begin{aligned}
    \text{Var}_{q(c|x)} {\boldsymbol \eta_c} & = \text{Tr} \big{(} \E_{q(c|x)} [ (\boldsymbol \eta_c - \E_{q(c|x)}\boldsymbol \eta_c)^T(\boldsymbol \eta_c - \E_{q(c|x)}\boldsymbol \eta_c) ] \big{)}, 
\end{aligned}
\end{equation*}
smaller.
\end{property}

Because of the convexity of partition function, the gradients of $\LL_d$ and $\text{Var}_{q(c|x)}\boldsymbol \eta_c$ are along the same direction,
\begin{equation*} \label{eq: gradient}
\begin{aligned}
     & ( \boldsymbol \nabla_{\boldsymbol \eta_c} \LL_d)^T  \cdot ( \boldsymbol  \nabla_{\boldsymbol \eta_c} \text{Var}_{q(c|x)} \boldsymbol \eta_c ) \sim \\
    & \left[  \nabla A|_{\eta_c} -  \nabla A|_{E_{q(c|x)}\eta_c}  \right]^T \cdot \left[\eta_c - (E_{q(c|x)}\eta_c) \right] \geq 0,
\end{aligned}
\end{equation*}
which means minimizing the dispersion term will implicitly minimize the  variance of prior parameters and makes these parameters tend to be the same.

Thus, according to the aforementioned two properties, the presence of dispersion term in the $\ELBO$ objective encourages the parameters of all mixture components to be as indistinguishable as possible and induces the mode-collapse.

In the following, we take Gaussian mixture and categorical mixture models as examples to show their specific dispersion term, and illustrate how their dispersion value are related to the degree of mode-collapse.

\noindent\textbf{$\LL_d$ in Gaussian Mixture VAE.}  We take the uni-variate Gaussian mixture VAE with known variance $\sigma$ as an example. The natural parameter of Gaussian distribution is $\eta=\mu / \sigma$, in which $\mu$ is known as the mean, and the log-partition function $A(\eta)=\eta^2/2$. As a result, the dispersion term could be written as $\frac{1}{2\sigma^2} [ \E_{q_\phi(c|x)}\mu_c^2 - (\E_{q_\phi(c|x)}\mu_c)^2 ],$
which is proportional to the variance of $\mu_c$.

\noindent\textbf{$\LL_d$ in Categorical Mixture VAE.} Considering the simplest categorical mixture VAE (CM-VAE), Bernoulli mixture VAE, the only natural parameter $\eta_c = \log \frac{p_c}{1-p_c}$ and the $A(\eta_c)=\log (1+e^{\eta_c})$. As a result, the dispersion term is $-\log (\prod_c p_c^{q_\phi(c|x)} + \prod_c (1-p_c)^{q_\phi(c|x)}).$

\begin{figure}[tp]
\footnotesize
\centering
    \subfloat[$\LL_d$ of GM-VAE]{
     \centering\label{fig:dispersion_gmvae}
     \includegraphics[scale=0.25]{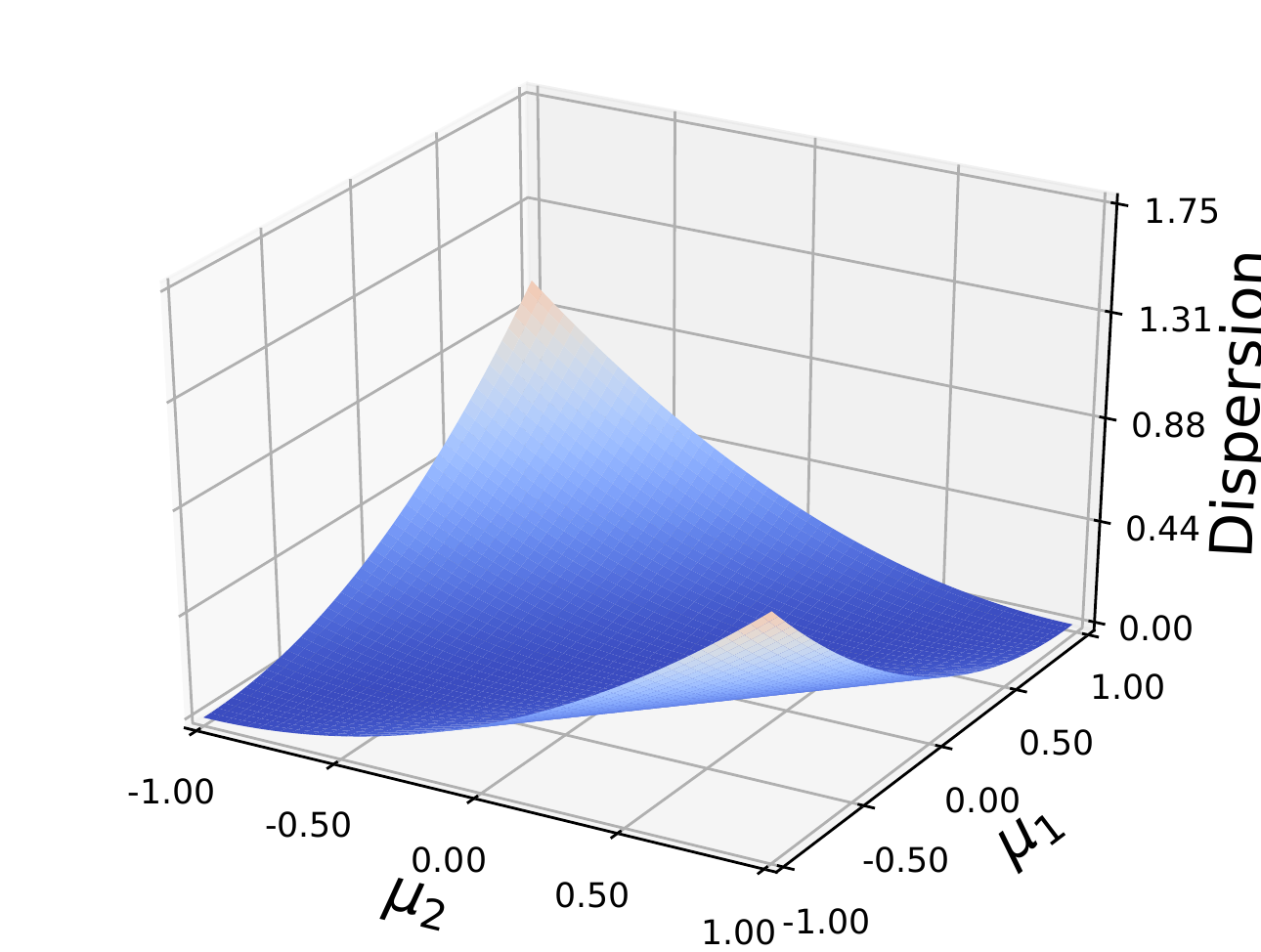}
      \hspace {3 pt}
     }
     \subfloat[$\LL_d$ of CM-VAE]{
      \hspace {3 pt}
     \centering\label{fig:dispersion_cmvae}
     \includegraphics[scale=0.25]{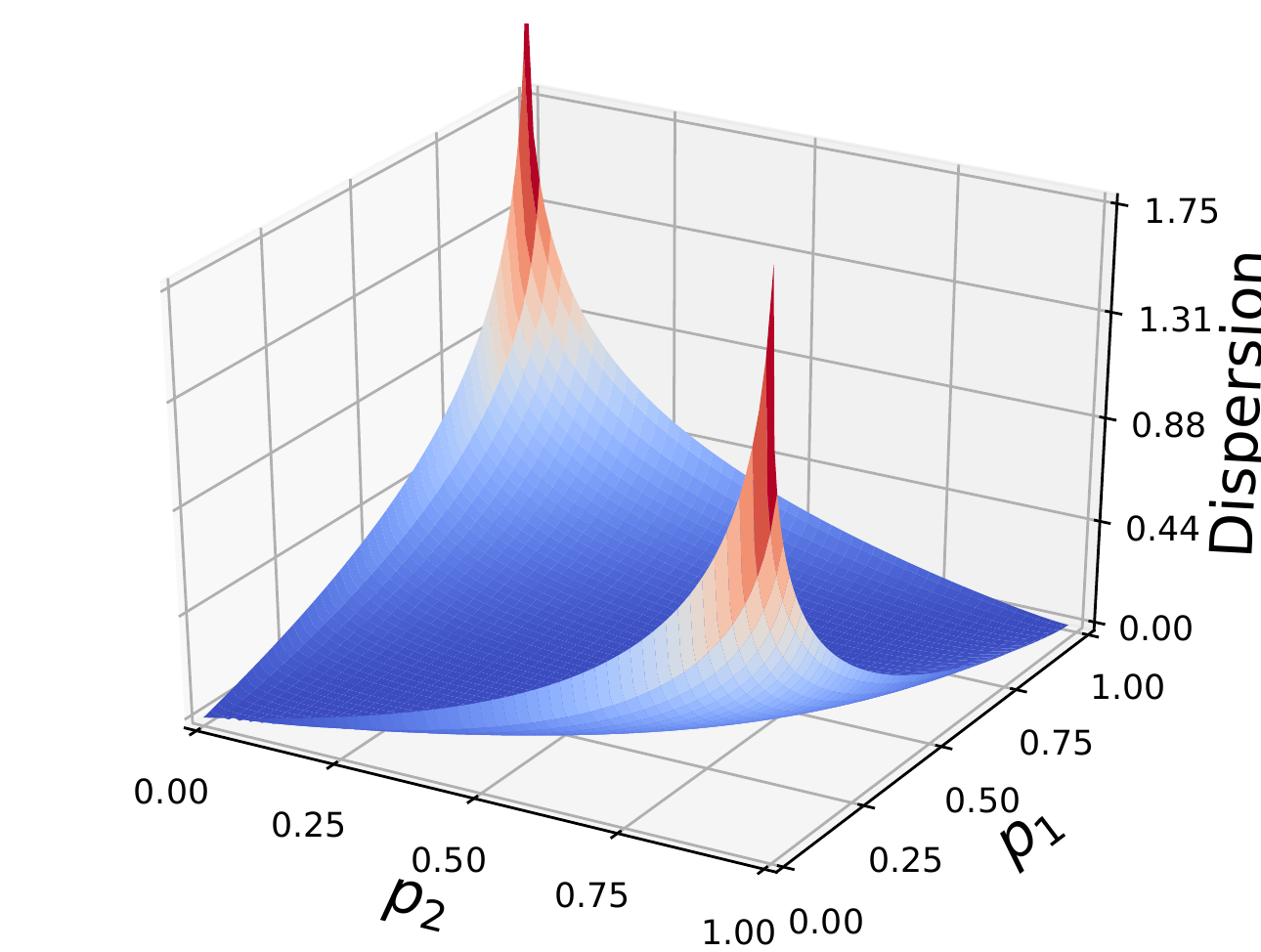}
       \hspace {3 pt}
     }
      \vspace{-6 pt}
    \caption{ Dispersion term $\LL_d$ of Gaussian and categorical (Bernoulli) mixture VAEs with respect to the parameters of two prior components.}\label{fig:dispersion}
\end{figure}

Visualization of these two dispersion term with respect to their prior parameters are shown in Fig.~\ref{fig:dispersion}. In Fig.~\ref{fig:dispersion_gmvae}, $\mu_1$ and $\mu_2$ are parameters of two components for one-dimensional GM-VAE, while in Fig.~\ref{fig:dispersion_cmvae} $p_1$ and $p_1$ are parameters of two components for CMVAE. $q_\phi(c|x)$ is assumed to be uniform. It is illustrated that the more similar the parameters are, the lower the dispersion item will be.

\subsection{Dispersed Exponential Family Mixture VAEs}
\label{sec:beta_gmvae}
In this section, we propose the Dispersed Exponential Family Mixture VAEs (DEM-VAE), which is a simple yet effective way to avoid the mode-collapse problem.
We takes the Gaussian mixture VAE (GM-VAE) as a specific example, and its corresponding proposed model is named as \methodshort.

According to the theoretical insights in Sec. \ref{sec:meancollapse}, we  include an extra positive  \textit{dispersion  term} to balance the mode collapse from $\ELBO$. 
The objective of DEM-VAE $\LL(\theta; x)$ for $x$ sampled from the dataset $D$ is:
\begin{equation}\label{eq:beta_gmvae}
\begin{aligned}
    & \LL(\theta; x) = \ELBO + \beta \cdot \LL_{d}
    , \\
    & \LL_d = \E_{q_\phi(c|x)}A(\boldsymbol \eta_c) - A(\E_{q_\phi(c|x)}\boldsymbol \eta_c). \nonumber
\end{aligned}
\end{equation}
For mixture of Gaussian, we have $\boldsymbol \eta_c = [\eta_1, \eta_2] = [\frac{\mu_c}{\sigma_c^2}, -\frac{1}{2\sigma_c^2}], A(\boldsymbol \eta_c)=- \frac{\eta_1^2}{4\eta_2} -\frac{1}{2} \log (-2\eta_2)$.
$\LL_d$ with a hyper-parameter $\beta$ is proposed to regularize the dispersion trends of mixture components. 
We can tune $\beta$ to make a trade-off between variance and concentration degree of prior components. 

The final objective of DEM-VAE could be:
\begin{equation}\label{eq:obj}
\begin{aligned}
    \frac{1}{|D|}\sum_{x\sim D} \Big{(} \E_{q_\phi(z|x)}\log p_\theta(x|z) - \KL(q_\phi(c|x)||p(c)) \\
    -\KL(q_\phi(z|x)||\hat{p}_{\overline{\eta}}(z|x))) - (1-\beta) \LL_d \Big{)}. 
\end{aligned}\nonumber
\end{equation}

\noindent\textbf{Additional Mutual Information Term}.
\quad As introduced by previous works \citep{chen2016infogan,zhao2017infovae,zhao2018unsupervised}, adding mutual information term to $\ELBO$ could enhance the interpretability and alleviate the KL-collapse.
Our method could be further improved in interpretability through adding the mutual information term 
\begin{equation}
\LL_{\text{mi}} = \HH(c) - \HH(c|x) = \E_{x} \E_{q_\phi(c|x)} (\log q_\phi(c|x) - \log q_\phi(c)) \nonumber
\end{equation}
to the objective. $q_\phi(c)$ could be estimated by $\E_x q_\phi(c|x)$ within mini-batch~\citep{zhao2018unsupervised}.

In the following, we take the Guassian mixture as an example and demonstrate the architecture of \methodshort for text generation.
Except for the learning objective, \methodshort has similar architecture as GM-VAE. It consists of a encoder for learning posterior and a decoder for generation. 

\noindent\textbf{Encoder}.
Recurrent neural networks such as GRU~\citep{chung2014empirical} as recognition networks encode sentences into compact hidden states. The parameters of posterior $q_\phi(z|x)$ and $q_\phi(c|x)$ are obtained based on hidden states. For example, mean $\mu_\phi$ and variance $\sigma_\phi^2$ of the posterior distribution $q_\phi(z|x)$~(assumed as a multivariate diagonal Gaussian) are obtained from two affine transformations. $q_\phi(c|x)$ could be modelled as a non-linear classifier taking the last hidden states as input.

\noindent\textbf{Decoder}.
In the decoding phase, we first sample a $z$ from a mixture priors~(in testing) or from posterior~(in training) by the reparameterization trick~\citep{kingma2013auto}.
The sentences will be generated on a recurrent neural language model fashion~(generation networks), with the $z$ as the initialized hidden state. $z$ could also be concatenated with embedding as input at each decoding step.

\noindent\textbf{Interpretable Dialog Generation}.
We follow the same approach of DI-VAE~\citep{zhao2018unsupervised} for interpretable dialog generation. 
The approach could be extended to other scenarios of interpretable generations, but we only validate our \methodshort on dialog for comparing with \citet{zhao2018unsupervised}.  

Specifically, in dialog generation, we generate response $r$ given the dialog context $y$. A \methodshort model is pre-trained in all utterances of the  training set to capture the interpretable facts~(component latent variable $c$) such as dialog actions or intentions.
In training, a hierarchical recurrent encoder-decoder model (HRED) with attention~\citep{sordoni2015hierarchical, serban2016building} $p_\theta(r|z,y)$ is trained to generate the response. Here $z$ is obtained from the pre-trained recognition network $q_\phi(z|r)$ of \methodshort and then fed into the decoder.
A policy network $p_\pi(c|y)$ is trained jointly to predict $c$ sampled from $q_\phi(c|r)$ in order to predict $c$  in the testing stage. 

\section{Experiments}
\label{sec:experiment}
\begin{figure*}[tb]  
\scriptsize
\centering
\begin{minipage}[t]{0.24\linewidth} 
    \centering 
 \subfloat[\tiny{GM-VAE \#2000}]{
 \label{fig:png_gmvae_2000}
 \centering
 \includegraphics[width=2.6 cm, height=2.0 cm]{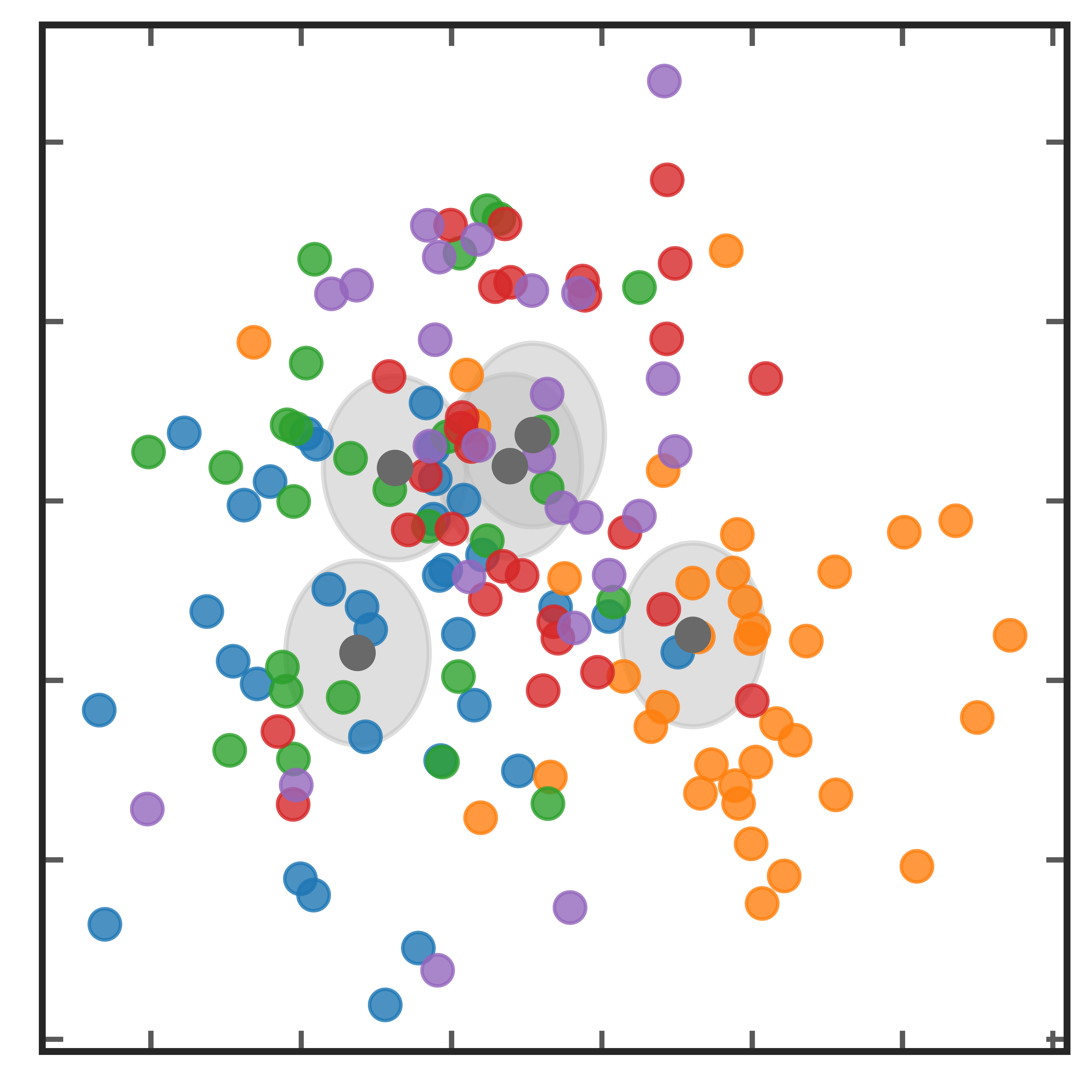}
 \hspace {4 pt}  
 }\\
 \vspace {-8 pt}
  \subfloat[\tiny GM-VAE \#10000]{
 \label{fig:png_gmvae_10000}
 \centering
 \includegraphics[width=2.6 cm, height=2.0 cm]{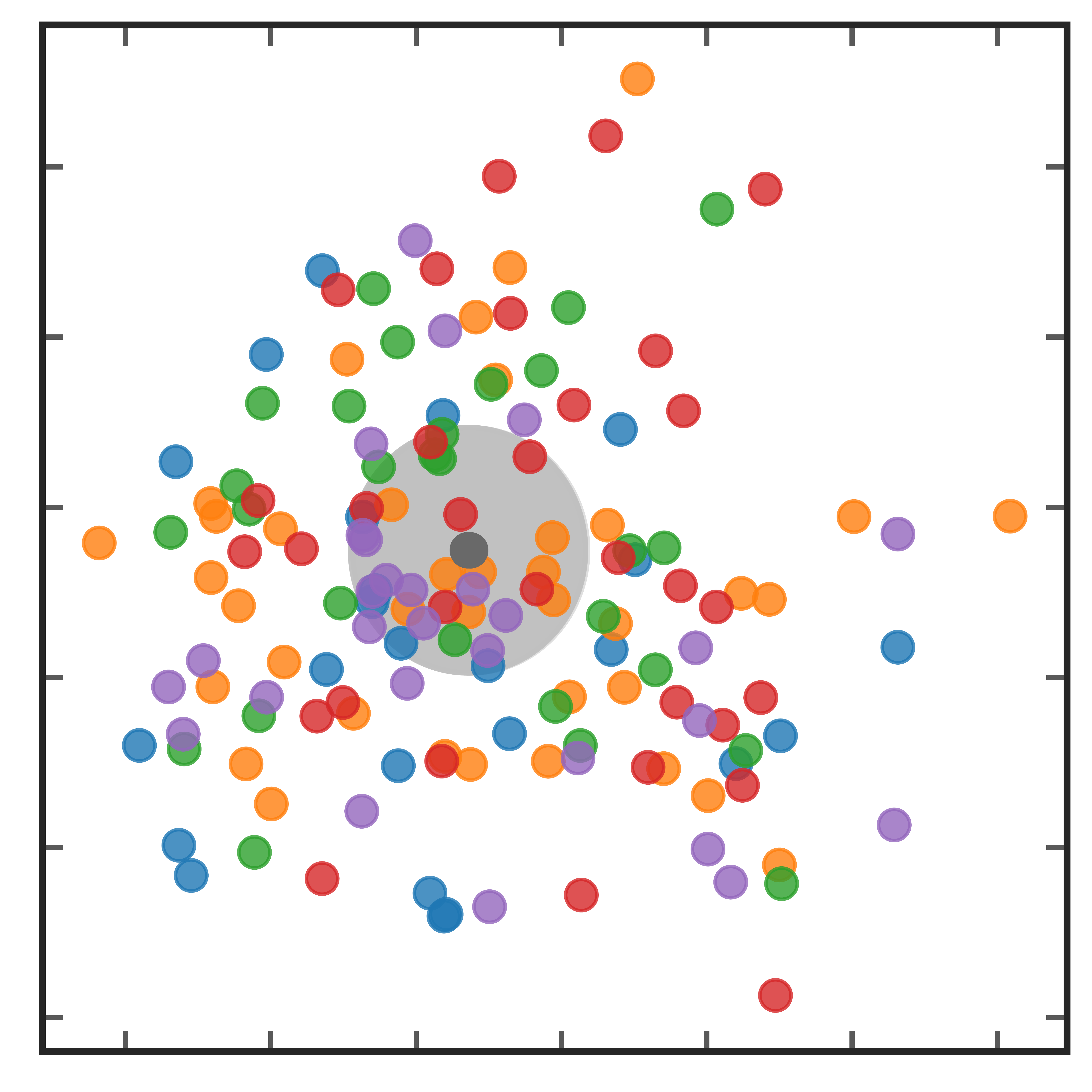}
 \hspace {4 pt}
 \vspace {-6 pt}
 }
  \end{minipage}%
  \begin{minipage}[t]{0.24\linewidth} 
    \centering 
      \subfloat[\tiny \dgmvae  \#2000]{
 \label{fig:png_woMI_2000}
 \centering
 \includegraphics[width=2.6 cm, height=2.0 cm]{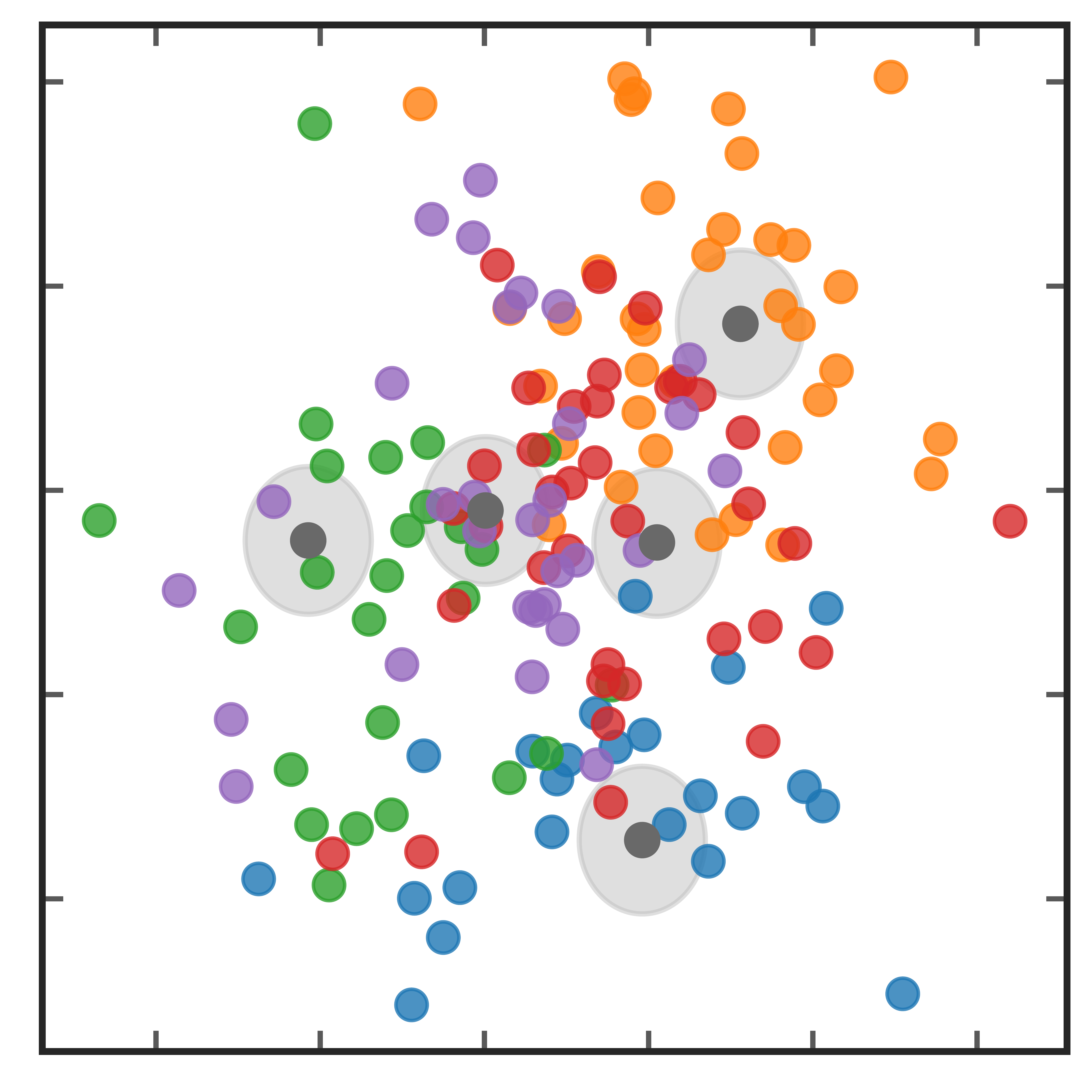}
 \hspace {4 pt}
 }\\
 \vspace {-8 pt}
  \subfloat[\tiny \dgmvae \#10000]{
 \label{fig:png_woMI_10000}
 \centering
 \includegraphics[width=2.6 cm, height=2.0 cm]{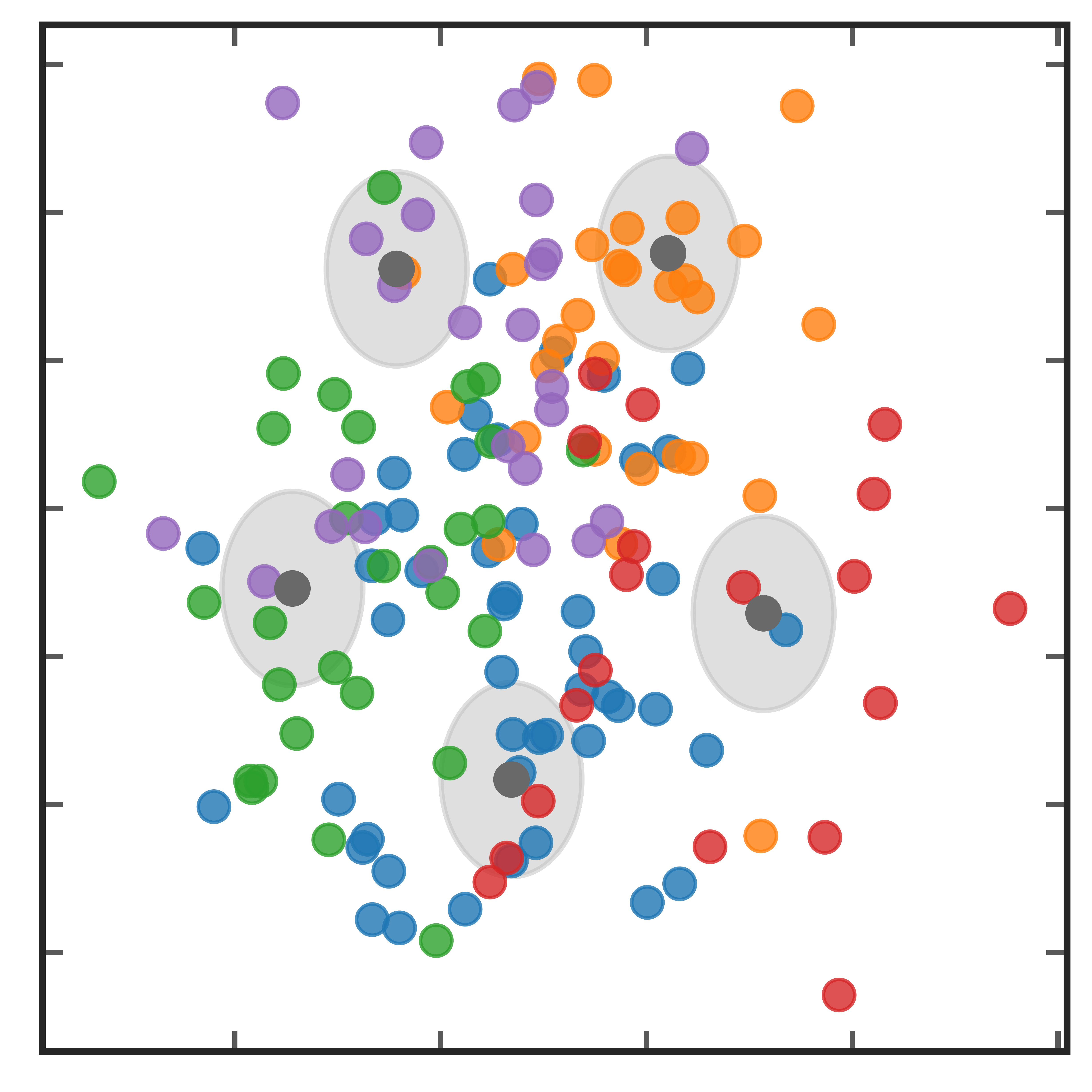}
 \hspace {4 pt}
 \vspace {-6 pt}
 }
\end{minipage}%
  \begin{minipage}[t]{0.24\linewidth} 
    \centering 
         \subfloat[\tiny GM-VAE $+$ $\mathcal L_{\text{mi}}$ \#2000]{
 \label{fig:png_woVM_2000}
 \centering
 \includegraphics[width=2.6 cm, height=2.0 cm]{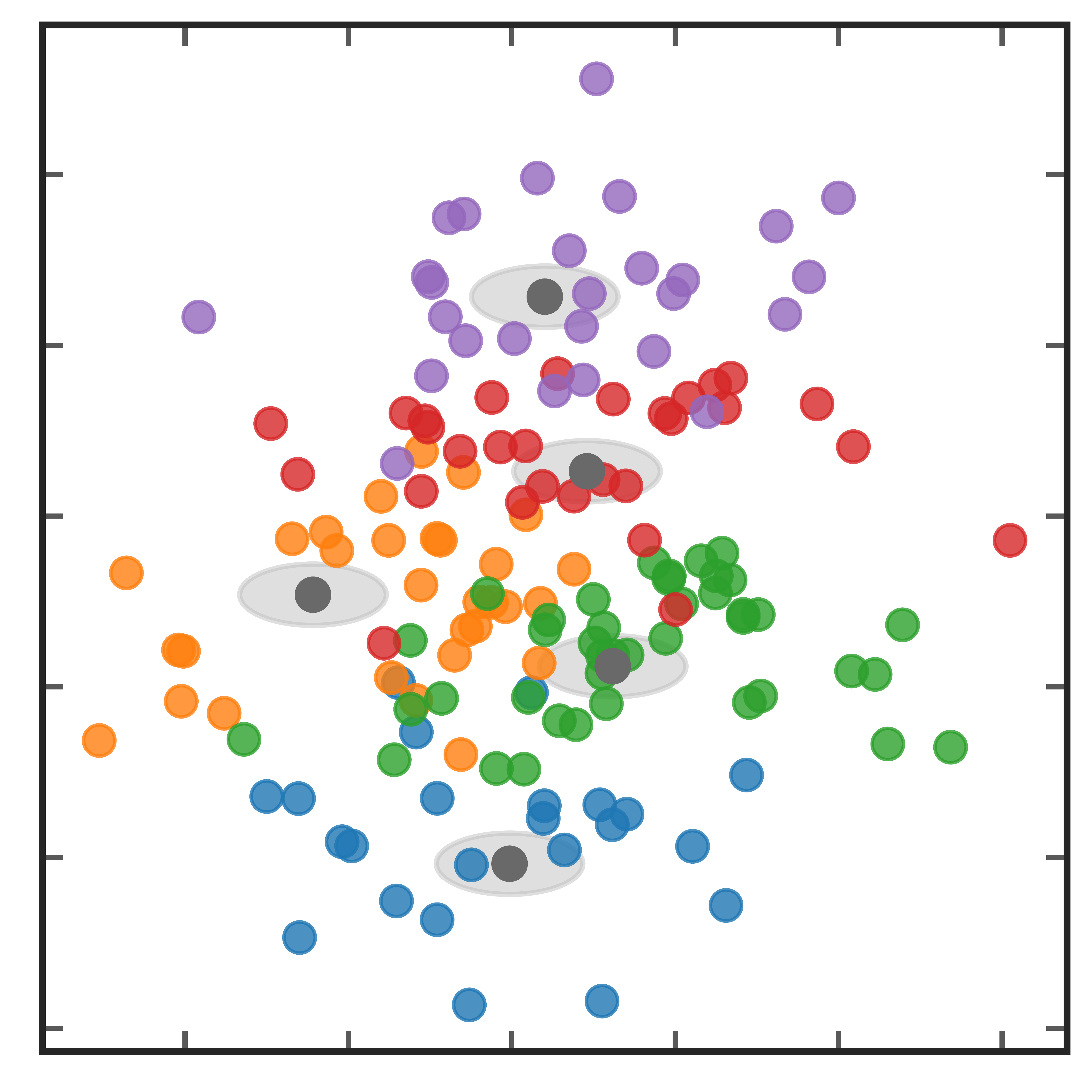}
 \hspace {4 pt}
 }\\
 \vspace {-8 pt}
  \subfloat[\tiny GM-VAE $+$ $\mathcal L_{\text{mi}}$ \#10000]{
 \label{fig:png_woVM_10000}
 \centering
 \includegraphics[width=2.6 cm, height=2.0 cm]{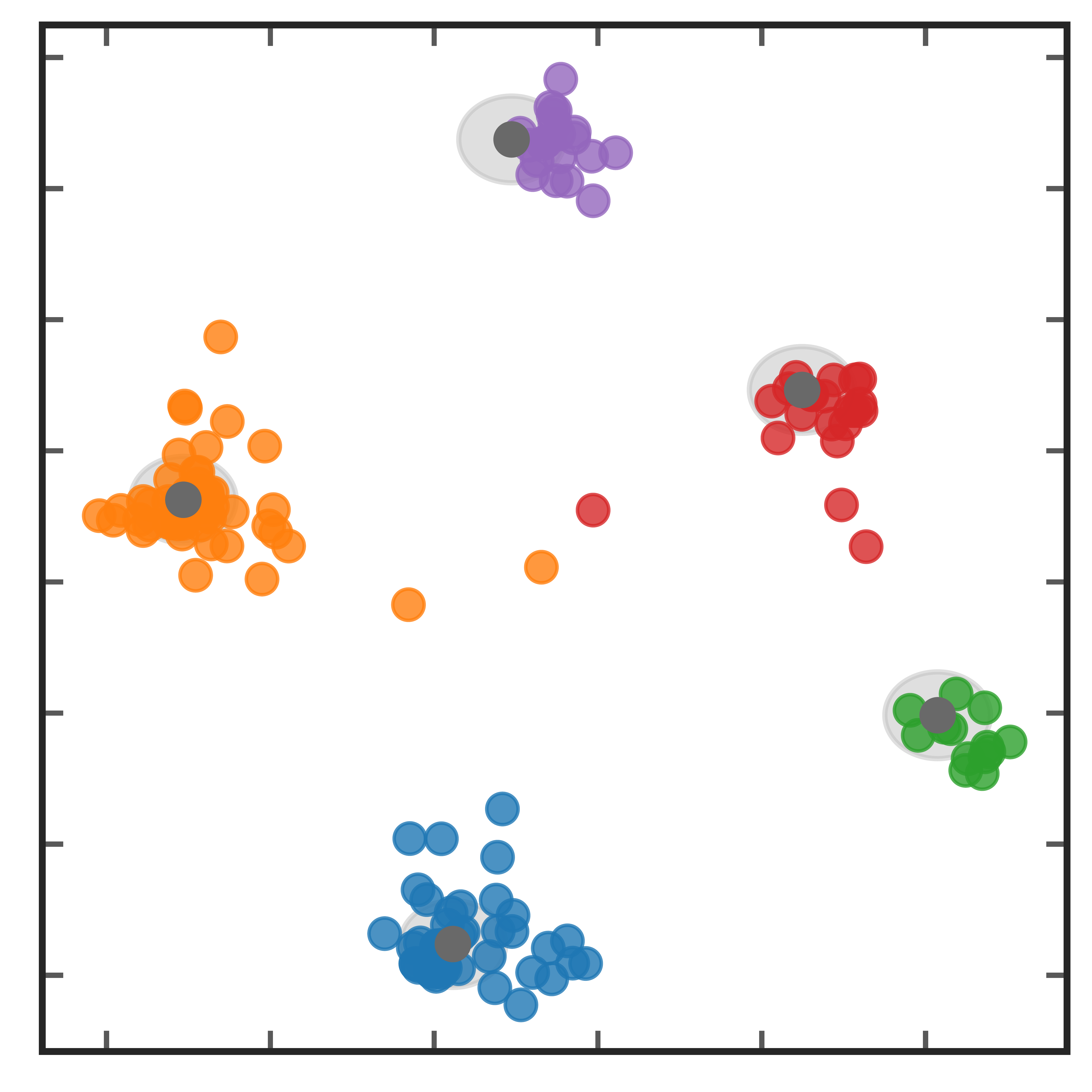}
 \hspace {4 pt}
 \vspace {-6 pt}
 }
\end{minipage}
\begin{minipage}[t]{0.24\linewidth} 
    \centering 
  \subfloat[\tiny \dgmvae $+$ $\mathcal L_{\text{mi}}$ \#2000]{
 \label{fig:png_betagmvae_2000}
 \centering
 \includegraphics[width=2.6 cm, height=2.0 cm]{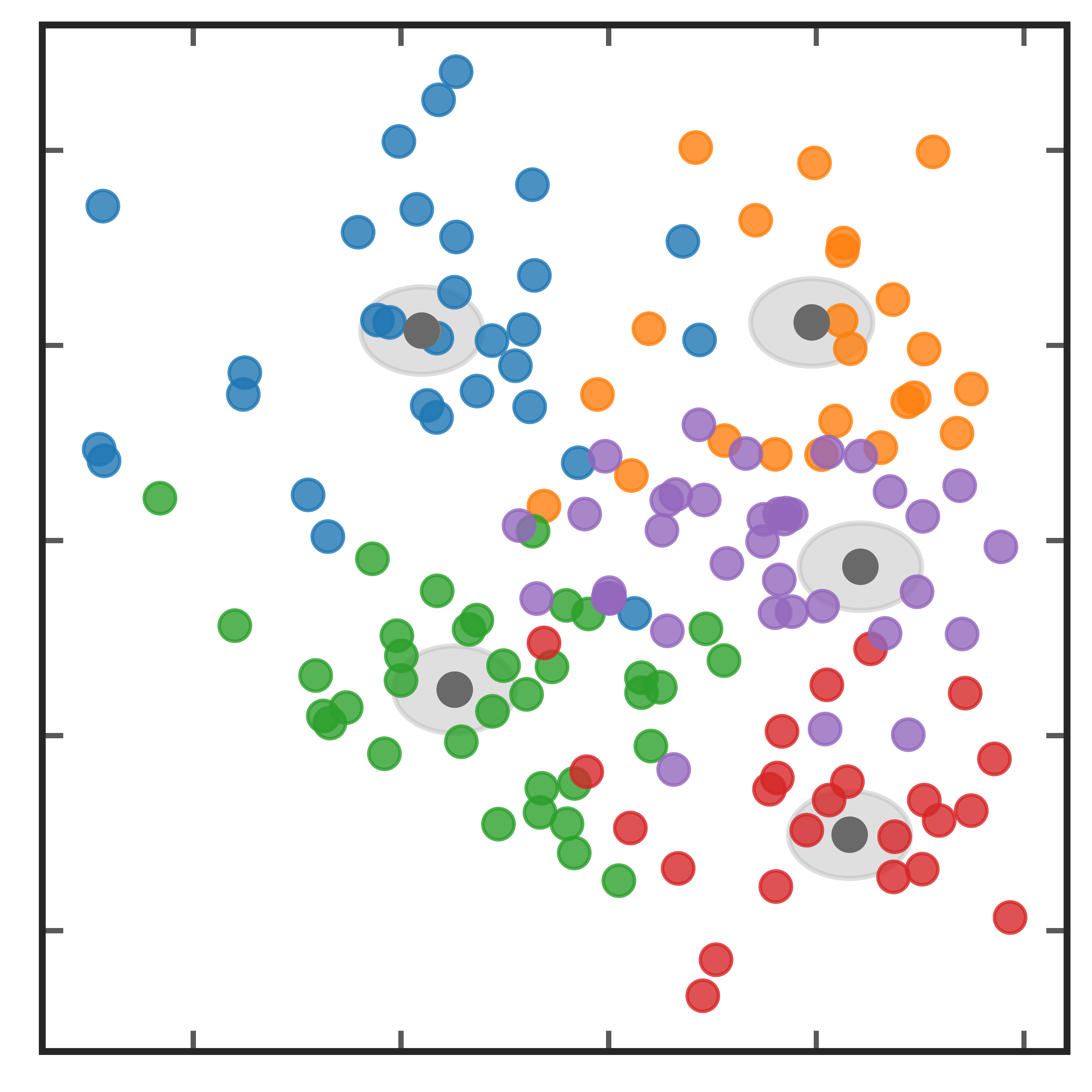}
 \hspace {4 pt}
 }\\
 \vspace {-8 pt}
  \subfloat[\tiny \dgmvae $+$ $\mathcal L_{\text{mi}}$ \#10000]{
 \label{fig:png_betagmvae_10000}
 \centering
 \includegraphics[width=2.6 cm, height=2.0 cm]{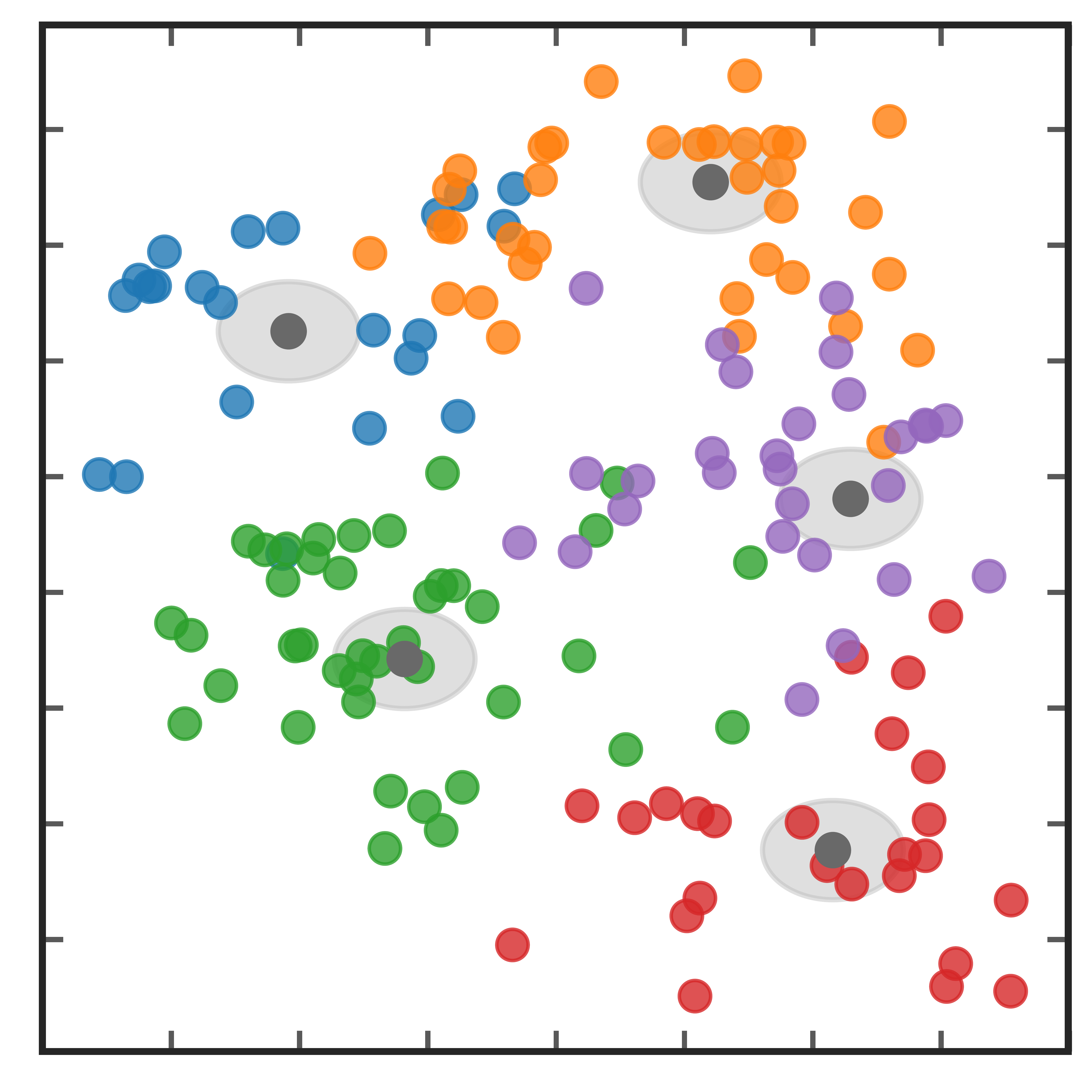}
 \hspace {4 pt}
 \vspace {-6 pt}
 }
  \end{minipage} 
 \caption{ Visualization of the mode collapse problem in DD dataset for GM-VAE, GM-VAE adding $\LL_{\text{mi}}$ term, \dgmvae and \dgmvae adding $\LL_{\text{mi}}$. Gaussian mixture priors are represented by grey points (mean) and circles (variance). The mean of posteriors are marked as colored points~(colors are associated with discrete latent variables).}
 \label{fig:mean_collapse_visualization}
\end{figure*}

In this section, we empirically test the generation quality and interpretable ability of two special cases of DEM-VAE, Dispersed GM-VAE~(\dgmvae) and Dispersed Multivariate Categorical Mixture VAE (DCM-VAE), on standard benchmarks. 
Results on dialog generation, attribute detection, and text classification demonstrate the superiority of \methodshort. 

\subsection{Setup}
\label{sec:setup}

We conduct experiments following \citet{zhao2018unsupervised}.
For generation quality, we use the Penn Treebanks~\citep[PTB]{marcus1993building} pre-processed by Mikolov~\citep{mikolov2010recurrent} as the benchmark. For interpretability, we use the Daily Dialogs~\citep[DD]{li2017dailydialog} and the Stanford
Multi-Domain Dialog~\citep[SMD]{eric2017key-value} datasets.
DD is a chat-oriented dataset containing 13,118 multi-turn dialogs, annotated with dialog actions and emotions. SMD contains 3,031 human-Woz, task-oriented dialogs collected
from 3 different domains~(navigation, weather and
scheduling).
In addition to unsupervised text generation tasks, we make use of emotion and action labels in DD as supervision and conduct supervised text generation experiments. 

We compare our model with the following baselines: 1) RNNLM, language model~\citep{mikolov2010recurrent} implemented by GRU~\citep{cho2014learning}; 2) AE, auto-encoders~\citep{Vincent2010Stacked} without latent space regularization; 
3) DAE,  auto-encoders with discrete latent space; 
4) VAE, the vanilla VAE~\citep{kingma2013auto} with only continuous latent variable and normal distribution prior; 5) DVAE, VAE with discrete latent variables; 6) DI-VAE, a DVAE variant~\citep{zhao2018unsupervised} adding extra mutual information term; 7) \textit{semi}-VAE, semi-supervised VAE model proposed by~\citet{kingma2014semi} with independent discrete and continuous latent variables; 8) GM-VAE, vanilla GM-VAE models as introduced in~\ref{sec:gmvae}.
Gumbel-softmax~\citep{jang2016categorical} is used for reparameterization in VAE variants with discrete latent variables.

The encoder and decoder in all models are implemented with single-layer  GRU~\citep{chung2014empirical}, with the hidden size as 512. For VAEs with discrete latent variables, multiple independent variables are adopted in order to increase model capacity. 
For unsupervised text generation, the dimension of discrete latent variables is set to 5 
while the number of discrete latent variables is set to 20, 3 and 3 for PTB, DD, and SMD. 
The total dimension of continuous latent space is set to 40 for PTB, 15 for DD and 48 for SMD. For supervised text generation, the discrete variable number is 30, the dimension of each variable is set to 8 and the number of mixture components is set to 30.
KL annealing with logistic weight function $1/(1+\exp(-0.0025(\text{step}-2500)))$ is adopted for all VAE variants. For GM-VAE, the KL annealing is applied in the whole KL term.
All hyper-parameters including $\beta$ are chosen according to the reverse perplexity (language generation task) or BLEU scores (dialog generation task) in the validation set. Details of hyper-parameters are included in the supplementary.

\subsection{Effects of \methodshort on Mode-Collapse}

We illustrate the effectiveness of \methodshort to alleviate the mode-collapse problem.
We train GM-VAE and \dgmvae in utterances on the DD dataset, and randomly sample 300 points from test data at 2,000 and 10,000 training steps, respectively. 
The dimension of latent space is set to 2 for visualization.
As in Fig.~\ref{fig:mean_collapse_visualization}, the mean and variance of each Gaussian component are indicated by grey points and circles, respectively. The means of posteriors are marked as colored points~(points with different discrete latent variables are associated with different colors).

It can be seen that, after 10,000 training steps, the vanilla GM-VAE degenerates into uni-Gaussian VAE, with the same mean values of all Gaussian components~(Fig.~\ref{fig:png_gmvae_2000} and \ref{fig:png_gmvae_10000}).
\dgmvae gives promising results as shown in Fig. \ref{fig:png_woMI_10000} and \ref{fig:png_betagmvae_10000}, in which different components of the GMM are dispersed and cluster data points into multiple modes well. 
As shown in Fig. \ref{fig:png_woVM_10000}, adding mutual information to GM-VAE indeed helps to alleviate the mode-collapse. However, the posterior points are quite concentrated to the priors, which makes the latent space degenerates into a discrete one so that the model cannot enjoy the high capacity and diversity of continuous variables.

\subsection{Language Generation Performance}

\begin{table*}[tb] 
\centering
\scriptsize
\begin{tabular}{l cccc cccc}
\toprule
 &  \multicolumn{4}{c}{\bf Evaluation Results} & \multicolumn{4}{c}{\bf Regularization Terms}  \\
\cmidrule(lr){2-5} \cmidrule(lr){6-9}
{\bf Model} & {\bf rPPL$^\downarrow$} & {\bf BLEU$^\uparrow$} & {\bf wKL$^\downarrow$} & {\bf NLL$^\downarrow$} & {\bf $\KL(z)$} & {\bf $\KL(c)$} & {\bf $\Var$} & {\bf $\LL_\text{mi}$} \\ 
\midrule
{ Test Set} & - & 100.0 & \textbf{0.14} & - & - & - & - & - \\ 
\midrule
{ RNN-LM~\citep{mikolov2010recurrent}} & -  & - & - & 101.21 & - & - & - & - \\
\midrule 
{ AE~\citep{Vincent2010Stacked}} & 730.81 & {\bf 10.88} & 0.58 & - & - & - & - & - \\ 
{ VAE ~\citep{kingma2013auto}} & 686.18 & 3.12 & 0.50 & $\leq$100.85 & 5.76 & - & - & - \\ 
\midrule
{ DAE } & 797.17 & 3.93 & 0.58 & - & - & - & - & - \\ 
{ \textsc{DVAE}} & 744.07 & 1.56 & 0.55  & $\leq$101.07 & - & 3.87 & - & 0.17 \\ 
{ \textsc{DI-VAE}~\citep{zhao2018unsupervised}} & 310.29 & 4.53 & 0.24 & $\leq$108.90 &  - & 24.78 & - & 1.14 \\ 
\midrule
{ \textit{semi}-VAE~\citep{kingma2014semi}} & 494.52 & 2.71 & 0.43 & $\leq$100.67  & 4.96 & 1.96 & - & 0.09 \\ 
{ \textit{semi}-VAE $+$ $\LL_{\text{mi}}$} & 260.28 & 5.08 & 0.20 & $\leq$107.30 & 0.045 & 24.04 & - & 1.12 \\ 
\midrule
{ \textsc{GM-VAE}} & 983.50 & 2.34 & 0.72 & \textbf{$\leq$99.44} & 3.41 & 0.00 & 0.001 & 0.00 \\ 
{ \textsc{GM-VAE} $+\LL_{\text{mi}}$} & 287.07 & 6.26 & 0.25 & $\leq$103.16 & 9.13 & 28.38 & 13.10 & 1.30 \\ 
\midrule
{ \dgmvae } & 257.68 & 8.17 & 0.19  & $\leq$104.26 & 41.48 & 4.76 & 787.03 & 0.22 \\ 
{ \dgmvae $+$ $\LL_{\text{mi}}$} & \textbf{247.37} & \textbf{8.67} & \textbf{0.18} & $\leq$105.73 & 25.48 & 19.73 &  203.34 & 0.91 \\ 
\bottomrule
\end{tabular}
\vspace{-6 pt}
\caption{ Language generation results on PTB. $\beta=0.8$ for \methodshort.  }\label{tab:generation_quality} 
\end{table*}

We evaluate the performance of language generation on PTB in Tab.~\ref{tab:generation_quality}, comparing \methodshort ($\beta=0.8$) with baselines described in Sec. \ref{sec:setup}. The test set of PTB is also included for comparison of text fluency.

We include four metrics to evaluate the generation performances: reverse perplexity (\textbf{rPPL}), \textbf{BLEU}~\citep{papineni2002bleu}, word-level KL divergence (\textbf{wKL}) and negative log-likelihood (\textbf{NLL}). Reverse perplexity is the perplexity of an LSTM language model~\citep{merity2017regularizing} trained on the synthetic data sampled from priors of VAE variants, and evaluated on the test set \citep{kim2017adversarially}. Lower rPPL indicates that the sampled sentences are more diverse and fluent. The BLEU score between input and reconstructed sentences reflects the ability of reconstruction. 
Word-level KL divergence between word frequencies calculated in generated data and training data shows the quality of generation. 
Negative log-likelihood\footnote{Evaluate by importance sampling~\citep{burda2015importance}. The number of sampling is 500.} reflects the generation ability of models.
These metrics are evaluated on the test set of PTB, except rPPL and wKL, which are calculated on sentences generated by sampling from these models' prior distribution\footnote{Sample size here is 40,000, the same as PTB training set.}~(sampling a random vector for AE).

Besides, the values of the regularization terms are also included in order to give some indications of the mode-collapse and KL collapse. We list the KL divergence of continuous latent variables (\textbf{KL(z)})
and discrete latent variables (\textbf{KL(c)}).
The weighted variance of parameters (\textbf{Var}\footnote{Var is calculated according to Eq.~\ref{eq:var} and taking the average over all samples. Variance of multiple parameters are summed.}) 
and mutual information (\textbf{MI}\footnote{MI is averaged over multivariate.}) terms are shown as well. 

We first present the ablation study to show the contribution of dispersion term and mutual information term.
Results are shown in Tab.~\ref{tab:generation_quality}. First of all, because extra bias is introduced in training objectives (higher mutual information or higher dispersion), the optimization object is no longer the lower bound of log-likelihood. As a result, the NLL will be worse with extra mutual information or dispersion term.
Adding an extra dispersion term helps to alleviate the mode-collapse, according to the higher variance value and mutual information compared with vanilla GM-VAE. Mutual information term helps to increase the information encoded by discrete latent variables, according to higher mutual information. 
Both terms improve the reconstruction performance and generation quality.

With the presence of both continuous and discrete latent variables, \dgmvae enjoys its higher model capacity and gives the best reconstruction performance~(BLEU), superior to other VAE variants.
Although \textit{semi}-VAE also includes discrete and continuous latent variables, it fails to make use of both of them because of the independent hypothesis. As shown in Tab.~\ref{tab:generation_quality}, either discrete or continuous latent variable collapses in \textit{semi}-VAEs.  AE could reproduce input sentences well, but it is not a true generative model and fails to generate diverse sentences.

Besides the reconstruction, we also find that \dgmvae outperforms related work in generating high-quality sentences.
{rPPL} is a powerful metric for measuring the fluency and diversity; \dgmvae obtains the lowest rPPL.
The lowest wKL also shows that word distribution in \dgmvae generations is most consistent with the training set.

\begin{figure}[t]
    \centering
    \includegraphics[scale=0.16]{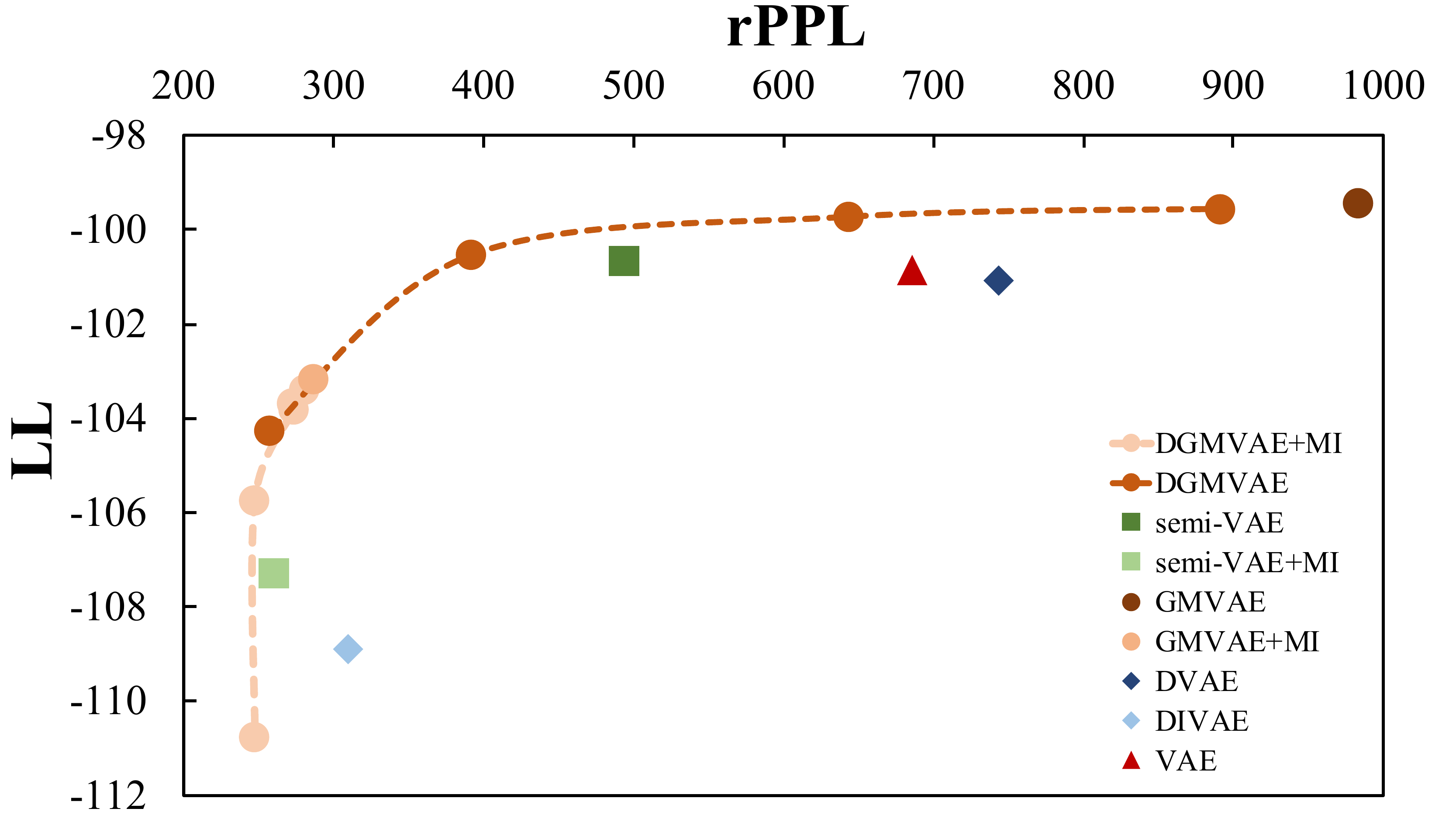}
    \vspace{-10 pt}
    \caption{LL and rPPL results for VAE models in PTB.}
    \label{fig:nll_rppl}
\end{figure}

Furthermore, we illustrate the superiority of Gaussian  mixture priors embodied by \methodshort.
The likelihood (LL) and reverse perplexity (rPPL) of \methodshort tuning $\beta$ compared with VAE baselines are shown in Fig.~\ref{fig:nll_rppl}.
Models, whose results are distributed closer to the upper left region, have better generation quality since it has relatively lower rPPL with the same LL. 
\methodshort is superior to other VAEs, because it has the optimal boundary. Semi-VAE is sub-optimal since it also contains both discrete and continuous latent variables, which is better than VAEs only include discrete or continuous latent variables.

\subsection{Interpretable Generation Results}
\label{sec:inter_gen_res}
In this section, the interpretable generation results of supervised, unsupervised and conditional text generation are shown. Following \citet{zhao2018unsupervised}, we include the experiments of interpretable language generation on DD and dialog generation on SMD.

\noindent\textbf{Unsupervised text generation: \dgmvae obtains the best performance in interpretability and reconstruction.}
In this experiment, we show the ability of \dgmvae to encode the dialog action and emotion information in latent variables unsupervisedly.
Because utterances in DD are annotated with Action and Emotion labels, we evaluate the ability of \dgmvae ($\beta=0.9$) to unsupervisedly capture these latent attributes on DD. We take the index $i$ with the largest posterior probability $q_\phi(c=i|x)$ as latent action labels. Following \citet{zhao2018unsupervised}, we use \textit{homogeneity} as the metric to evaluate the consistency between golden action and emotion labels with labels obtained from \dgmvae. The number of our labels is 125. 
Results of homogeneity of action~(act) and emotion~(em) together with MI term and BLEU are shown in Tab. \ref{tab:dd}.
It shows that \dgmvae outperforms other VAEs in reconstruction and gives the best homogeneity on both the action and emotion.

\begin{table}[tb]
    \scriptsize
    \centering
    \hspace{-2 pt}
     \begin{tabular}{l c c c c}
    \toprule
     &   \multicolumn{4}{c}{\bf Unsupervised DD}   \\
     \cmidrule(lr){2-5}
    {\bf Model}  & {\bf MI} & {\bf BLEU$^\uparrow$} & {\bf act$^\uparrow$} & {\bf em$^\uparrow$} \\
    \midrule
    { \text{DI-VAE}} &   1.20 & 3.05 & 0.18 & 0.09\\
    \midrule
    { \textit{semi}-VAE} &  0.03 & 4.06 & 0.02 & 0.08\\ 
    { \textit{semi}-VAE $+$ $\LL_{\text{mi}}$} &   1.21 & 3.69 & 0.21 & 0.14\\ 
    \midrule
    { \textsc{GM-VAE}} &   0.00 & 2.03 & 0.08 & 0.02\\
    { \textsc{GM-VAE} $+$ $\LL_{\text{mi}}$}  &  1.41 & 2.96 & 0.19 & 0.09\\
    \midrule
    { \dgmvae } &   0.53 & {\bf 7.63} & 0.11 & 0.09\\ 
    { \dgmvae $+$ $\LL_{\text{mi}}$} & 1.32 & 7.39 & {\bf 0.23}  & {\bf 0.16} \\
    \bottomrule
    \end{tabular}
    \vspace{-6 pt}
     \caption{ Results of interpretable language generation on DD. Mutual information (MI), BLEU and homogeneity with actions (act) and emotions (em) are shown.  The larger$^\uparrow$, the better. 
     }\label{tab:dd}
\end{table}

\begin{table}[tb]
    \scriptsize
    \centering
     \begin{tabular}{lllll}
    \toprule
    & \multicolumn{4}{c}{\bf Automatic Metrics}  \\ 
    \cmidrule(lr){2-5}
    {\bf Model} & {\bf BLEU} & {\bf Ave.} & {\bf Ext.} & {\bf Grd.}  \\
    DI-VAE & 7.06 & 76.17 & 43.98 & 60.92  \\
    \dgmvae $+\LL_{\text{mi}}$ & \bf 10.16 & \bf 78.93 & \bf \bf 48.14 & \bf 64.87  \\
    \midrule
     & \multicolumn{4}{c}{\bf Human Evaluation} \\
     \cmidrule(lr){2-5}
    {\bf Model} & \multicolumn{2}{c}{\bf Quality} & \multicolumn{2}{c}{\bf Consistency} \\
    DI-VAE & \multicolumn{2}{c}{2.31} & \multicolumn{2}{c}{3.08} \\
    \dgmvae $+\LL_{\text{mi}}$  & \multicolumn{2}{c}{\bf 2.45} & \multicolumn{2}{c}{\bf 3.35} \\
    \bottomrule
    \end{tabular}
    \vspace{-6 pt}
     \caption{ Dialog evaluation results on SMD. Four automatic metrics: BLEU, average (Ave.), extrema (Ext.) and greedy (Grd.) word embedding based similarity are shown. 
     Response quality and consistency within the same $c$ are scored by human. }\label{tab:smd_quality}
\end{table}

\noindent\textbf{Conditional text generation: \dgmvae gives better dialog generation quality and interpretability.} 
Furthermore, we evaluate the ability of interpretable dialog generation of \methodshort based on the structured latent space on SMD dataset. Both automatic evaluation and human evaluation are conducted. BLEU and three word embedding\footnote{We use GloVe~\citep{pennington2014glove} word embeddings of 300 dimension trained on 840B tokens from https://nlp.stanford.edu/projects/glove/.} based topic similarity~\citep{serban2017a}: Embedding Average, Embedding Extrema and Embedding Greedy~\citep{mitchell2008vector, forgues2014bootstrapping, rus2012comparison} are used to evaluate the quality of responses. 
In addition, three human evaluators were asked to score the quality (from 0 to 3) of 159 responses generated by DI-VAE and \dgmvae. Because SMD does not offer human-annotated action labels of dialog utterances, we follow~\citet{zhao2018unsupervised} to label dialog actions by human experts for each discrete latent variable $c$, according to their sampled utterances. Another 3 annotators are asked to evaluate the consistency between the action name and another 5 sampled utterances, showing the interpretability.

Results are shown in Tab.~\ref{tab:smd_quality}. Both automatic and human evaluations show that \dgmvae obtains better generation quality and interpretability than DI-VAE on SMD. 
We perform one-tail t-tests on human evaluation scores and find that the superiority of our model is significant in both quality and consistency with p-values no more than 0.05.

\begin{table}[tb]
    \centering
    \scriptsize
    \begin{tabular}{ll}
    \toprule
    {\bf Action} & Inform-route/address \\
    \midrule
    \multirow{3}{*}{\bf Utterance} & There is a Safeway 4 miles away. \\
    & There are no hospitals within 2 miles. \\
    & There is Jing Jing and PF Changs. \\
    \bottomrule
    \toprule
    {\bf Action} & Request-weather \\
    \midrule
    \multirow{3}{*}{\bf Utterance} & What is the weather today? \\
    & What is the weather like in the city? \\
    & What's the weather forecast in New York? \\
    \bottomrule
    \end{tabular}
    \vspace{-6 pt}
    \caption{ Example actions (Act) and corresponding utterances (Utt) discovered by \dgmvae on SMD. The action name is annotated by experts. }\label{tab:case-actions} 
\end{table}

\begin{table}[t]
\centering
\scriptsize
\begin{tabular}{ll}
\toprule
{\bf Context} & \textit{Sys:} Taking you to Chevron. \\
\midrule
\multirow{2}{*}{\bf Predict} & (1-1-3, thanks) Thank you car, let's go there! \\
& (1-0-2, request-address) What is the address? \\
\bottomrule
\toprule
{\bf Context} & \textit{User:} Make an appointment for the doctor. \\
\midrule
\multirow{4}{*}{\bf Predict} & (3-2-4, set-reminder) Setting a reminder for \\ 
& your doctor's appointment on the 12th at 3pm.  \\
& (3-0-4, request-time) What time would you   \\
& like to be schedule your doctor's appointment? \\
\bottomrule
\end{tabular}
\vspace{-6 pt}
\caption{ Dialog cases on SMD, which are generated by sampling different $c$ from policy network. The label of sampled $c$ are listed in parentheses with the annotated action name.}\label{tab:case-response}
\end{table}

We perform case studies to validate the performance of \dgmvae  qualitatively.
Some dialog actions with their utterances discovered by \dgmvae are shown in Tab.~\ref{tab:case-actions}. 
It can be seen that utterances of the same actions could be assigned with the same discrete latent variable $c$. 
We also give some dialog cases generated by \dgmvae in Tab. \ref{tab:case-response} with their contexts. 
Given the same context, responses with different actions are generated by sampling different values of discrete latent variables, which shows that \dgmvae has the ability to generate diverse and interpretable responses.
More cases can be found in the supplementary materials.

\begin{table}[t] 
\centering
\scriptsize
\begin{tabular}{lcccc}
    \toprule
     &   \multicolumn{4}{c}{\bf Supervised DD}   \\
     \cmidrule(lr){2-5}
    {\bf Model}  & {\bf NLL$^\downarrow$} & {\bf ACC$_\text{act}$ $^\uparrow$} & {\bf ACC$_\text{em}$ $^\uparrow$} & {\bf $\Var$} \\
    \midrule
    { \text{DVAE}} &   $\leq$\textbf{48.71} &  0.79 & 0.63 & -\\ 
    { \textsc{CM-VAE}} & $\leq$52.24 &   0.79 & 0.63 & 0.00\\ 
    { \dcmvae}  & $\leq$48.78 & \textbf{0.80} & \textbf{0.73} & 139.16\\ 
    \bottomrule
\end{tabular}
\vspace{-6 pt}
\caption{Results of supervised interpretable generation on DD. Negative log-likelihood (NLL), classification accuracy (ACC) on emotion (em) and action (act) and the variance of parameters ($\Var$) are shown.}
\label{tab:sup_dd}
\end{table} 

\noindent\textbf{Supervised text generation: \dcmvae obtains best performance in multi-label classification accuracy and reconstruction.}
\quad
To show that mode-collapse is a general problem for exponential family mixture VAEs and our proposal is a general solution for it, we conducted supervised text generation experiments using mixture of categorical VAE (CM-VAE). CM-VAE includes mixture of multivariate categorical distribution as priors, which is able to capture the dependency between multiple variables and has been used in multi-label classification in image~\citep{li2016conditional}.

The supervised generation experiment is conducted in DD. Each utterance in DD has emotion and action labels.
Multivariate categorical distributions are taken as priors,  two of whose components are related to emotion and action through adding cross entropy with golden labels to the objective. 

Experimental results of baselines and \dcmvae ($\beta=0.4$) are shown in Tab.~\ref{tab:sup_dd}. The classification accuracy of action and emotion is calculated by the corresponding posterior network.
\dcmvae could alleviate the mode-collapse effectively and obtains the best performance in multi-label classification accuracy with slightly worse NLL. It demonstrates that under supervision, \dcmvae can inject the interpretable properties in latent variables well.
Mode collapse occurs in vanilla \textsc{CM-VAE}, making its performance not better than baseline. \dcmvae is able to make full use of multiple modes and capture the dependency of variables, which improves the performance of multi-label classification.

\subsection{Comparison with KL Collapse Solutions}

In this section, we compare our method with previous solutions of fixing the KL collapse problem in VAEs, and show the  superiority of DEM-VAE on solving the specific mode-collapse problem in EM-VAEs.

Taking Guassian Mixture VAEs as an example, we compare our \methodshort with following baselines: 1) $\beta$-GM-VAE~\citep{higgins2017beta-vae}: re-weighting the whole KL term $\KL(q_\phi(z,c)||p_\eta(z,c))$ by $\beta$; 2) aggressive update~\cite{he2018lagging}: aggressively updating the parameters of posterior network to alleviate posterior collapse; 3) free bits (FB)~\citep{kingma2016improved, li2019surprisingly}: constrain the values of $\KL(c)$ no larger than $\lambda_c$ and $\KL(z)$ no larger than $\lambda_z$. In our experiment, $\lambda_c=5$ and $\lambda_z=10$ are chosen. 
The same logistic annealing method is applied in KL terms. For $\beta$-GM-VAE, we test the performance on a large $\beta=0.9$, a medium $\beta=0.5$ and a small $\beta=0.1$. 

Results on PTB are shown in 
Tab.~\ref{tab:kl_collapse}. Among these methods, \dgmvae is the most effective way to avoid mode-collapse. In the following, we compare with each of these methods.

GM-VAE with annealing fails to avoid mode-collapse. Because KL-annealing is not specific for solving the mode-collapse problem, it can prevent the $\KL(z)$ from collapsing to zero but fails to prevent the mode-collapse. 

Aggressive updating also fails to fix the mode-collapse. In the early aggressive updating stage, it could keep the $\KL(c)$ term in the non-zero level with $\KL(c)\sim 8.0$ but is not well-learned with $\ELBO\sim 240$. However, mode-collapse still occurs for the subsequent standard VAE training stage. Because no explicit interference on priors is introduced, more updates for posterior network is difficult to guide the priors to be dispersed.

FB can also avoid mode collapse to some extent, but the constraint on the whole $\KL(z)$ term still prefers to make the variance of parameters to be small. Our method could be combined with the FB by replacing the re-weighting of dispersion term with a maximum value constraint.

$\beta$-GM-VAE adjusts the weight of the whole KL term, but it is hard to find a suitable $\beta$ to maintain a good generative quality while having a non mode-collapse latent space.
When $\beta$ is small (0.1), the NLL is high because weakening the whole KL has a negative effect of under-regularization. When $\beta$ is relative high (0.5, 0.9), it fails to embed enough information in discrete latent variables as expected. 

Our proposed dispersion term is specifically designed for the mode-collapse problem, it is easier to learn a solution with both good generative quality and informative discrete latent variables.

\begin{table}[t] 
\centering
\scriptsize
\begin{tabular}{lcccccc}
    \toprule
    {\bf Model}  & {\bf NLL$^\downarrow$} & {\bf BLEU$^\uparrow$} & {\bf $\LL_\text{mi}$} & {\bf $\Var$} & {\bf $\KL(c)$} & {\bf $\KL(z)$} \\
    \midrule
    \textsc{GM-VAE} &  99.44 & 2.34 & 0.00 & 0.001 & 0.00 & 3.41  \\
    \quad + aggressive &  99.64 & 1.47 & 0.00 & 0.00 & 0.00 & 2.35 \\
    \quad + FB  & 101.39 & 6.59 & 0.19 & 16.85 & 5.14 & 12.52 \\
    \quad w/ $\beta=0.1$ &  136.08 & 5.88 & 0.15 & 14.89 & 4.11 & 49.49 \\
    \quad w/ $\beta=0.5$ &  103.32  & 5.92 & 0.024 & 1.57 & 1.43 & 21.12 \\
    \quad w/ $\beta=0.9$ &  100.05 & 1.78 & 0.00 & 0.10 & 0.44 & 1.52 \\
    \midrule
    { \dgmvae}  & 104.26 & 8.17 & 0.22 & 787.03 & 4.76 & 41.48 \\ 
    \bottomrule
\end{tabular}
\vspace{-6 pt}
\caption{Results of \dgmvae and methods of fixing the posterior collapse. The NLL in the table is a upper-bound estimated by importance sampling.}
\label{tab:kl_collapse}
\end{table}

\section{Conclusion}
\label{sec:conclusion}
In this paper, we introduce a family of models for text generation using VAEs with exponential family mixture as priors. 
We theoretical analyze the cause of mode-collapse problem in the training.
We propose a method to fix the problem and effectively train the models. 
Experimental results show that our method achieves good performance in interpretable text generation.

\section*{Acknowledgements}

We thank the reviewers for providing valuable comments to improve this paper. We thank Shenjian Zhao and Jiangtao Feng for helpful feedback.




\bibliography{iclr2020_conference}
\bibliographystyle{icml2020}


\end{document}